\documentclass[dvipsnames]{article} %
\usepackage{colm2024_conference}
\usepackage{threeparttable}
\usepackage{booktabs}
\usepackage{graphicx}
\usepackage{enumitem}
\usepackage{wrapfig}
\usepackage{algorithm}
\usepackage{algpseudocode}

\usepackage{microtype}
\usepackage{amsmath}
\usepackage{colortbl}
\usepackage[utf8]{inputenc}
\definecolor{lightgray}{rgb}{0.9,0.9,0.9}
\usepackage{fancyhdr}
\usepackage{caption}
\usepackage{subcaption}
\usepackage{setspace}
\usepackage{url}
\usepackage{multirow}
\usepackage{colortbl}
\usepackage{tabularx}
\usepackage{blindtext}
\usepackage{pgfplots}
\pgfplotsset{compat=1.18} 
\usepackage{tikz}
\usetikzlibrary{er,positioning,bayesnet}
\usepackage{makecell}
\usepackage{tipa}
\usepackage{siunitx}
\usepackage{nicefrac}
\usepackage{tocloft}
\usepackage{listings}
\usepackage[raster,skins]{tcolorbox} %
\usepackage{xltabular}
\usepackage{adjustbox}
\usepackage{xurl}
\usepackage{rotating}
\usepackage[normalem]{ulem}
\useunder{\uline}{\ul}{}
\usepackage{cleveref}
\Crefname{figure}{Figure}{Figures}
\Crefname{table}{Table}{Tables}
\Crefname{equation}{Equation}{Equations}
\usepackage{CJKutf8}
\usepackage{titlesec}
\usepackage{wrapfig}
\usepackage{verbatim}
\tcbuselibrary{listingsutf8}
\usepackage{titletoc}
\usepackage{adjustbox}
\usepackage{titling}

\setlist[itemize]{left=4pt}
\setlist[enumerate]{left=4pt}
\definecolor{lightgray}{rgb}{0.9,0.9,0.9} % You can adjust the RGB values for different shades

\definecolor{citecolor}{HTML}{4d5eff}
\definecolor{tabblue}{RGB}{31,119,180}
\definecolor{taborange}{RGB}{255,127,14}
\definecolor{tabgreen}{RGB}{44,160,44}
\definecolor{darkblue}{rgb}{0, 0, 0.5}

\newtcolorbox{AIbox}[2][]{aibox,title=#2,#1}
\definecolor{lightblue}{rgb}{0.22,0.45,0.70}%
\definecolor{Gray}{gray}{0.95}
\definecolor{Cornsilk}{rgb}{1.0, 0.97, 0.86}

\usepackage{amsmath,amsfonts,bm}

\def\eqref#1{equation~\ref{#1}}
\def\1{\bm{1}}

\DeclareMathAlphabet{\mathsfit}{\encodingdefault}{\sfdefault}{m}{sl}
\SetMathAlphabet{\mathsfit}{bold}{\encodingdefault}{\sfdefault}{bx}{n}

\renewcommand{\hflink}{https://huggingface.co/Skywork/Skywork-R1V3-38B}
\renewcommand{\ghlink}{https://github.com/SkyworkAI/Skywork-R1V}

\newcommand*\justify{%
  \fontdimen2\font=0.4em% interword space
  \fontdimen3\font=0.2em% interword stretch
  \fontdimen4\font=0.1em% interword shrink
  \fontdimen7\font=0.1em% extra space
  \hyphenchar\font=`\-% allowing hyphenation
}

\renewcommand{\texttt}[1]{%
  \begingroup
  \ttfamily
  \begingroup\lccode`~=`/\lowercase{\endgroup\def~}{/\discretionary{}{}{}}%
  \begingroup\lccode`~=`[\lowercase{\endgroup\def~}{[\discretionary{}{}{}}%
  \begingroup\lccode`~=`.\lowercase{\endgroup\def~}{.\discretionary{}{}{}}%
  \catcode`/=\active\catcode`[=\active\catcode`.=\active
  \justify\scantokens{#1\noexpand}%
  \endgroup
}

\newtcblisting{markdownbox}{
    colback=blue!5,        % light blue background
    colframe=blue!75!black, % darker blue frame
    listing only,
    boxrule=0.8mm,
    arc=2mm,
    outer arc=2mm,
    left=3mm,
    right=3mm,
    top=2mm,
    bottom=2mm,
    width=\linewidth,
    fonttitle=\bfseries,
    title=Annotation Protocol,
    breakable               % allow content to break across pages
}

\newtcblisting{markdownboxpa}{
    colback=blue!5,        % light blue background
    colframe=blue!75!black, % darker blue frame
    listing only,
    boxrule=0.8mm,
    arc=2mm,
    outer arc=2mm,
    left=3mm,
    right=3mm,
    top=2mm,
    bottom=2mm,
    width=\linewidth,
    fonttitle=\bfseries,
    title=Prompt for Preference Attribute Generation,
    breakable               % allow content to break across pages
}

\newtcblisting{markdownboxannotation}{
    colback=blue!5,        % light blue background
    colframe=blue!75!black, % darker blue frame
    listing only,
    boxrule=0.8mm,
    arc=2mm,
    outer arc=2mm,
    left=3mm,
    right=3mm,
    top=2mm,
    bottom=2mm,
    width=\linewidth,
    fonttitle=\bfseries,
    title=Prompt for Preference Pair Annotation,
    breakable               % allow content to break across pages
}

\newtcblisting{markdownboxexamples}{
    colback=teal!5,         % light teal background
    colframe=teal!60!black, % darker teal frame
    listing only,
    boxrule=0.8mm,
    arc=2mm,
    outer arc=2mm,
    left=3mm,
    right=3mm,
    top=2mm,
    bottom=2mm,
    width=\linewidth,
    fonttitle=\bfseries,
    title=Example of Preference Attributes,
    breakable               % allow content to break across pages
}

\title{Skywork-R1V3 Technical Report}

\author{
 Multimodal Team \\[0.3em]
\textbf{Skywork AI, Kunlun Inc}   \\[0.3em]
}

\begin{document}

\maketitle

\begin{abstract}
We introduce Skywork-R1V3, an advanced, open-source vision-language model (VLM) that pioneers a new approach to visual reasoning. Its key innovation lies in effectively transferring reasoning skills from text-only Large Language Models (LLMs) to visual tasks.  The strong performance of Skywork-R1V3 primarily stems from our elaborate post-training RL framework, which effectively activates and enhances the model’s reasoning ability, without the need for additional continue pre-training. Through this framework, we further uncover the fundamental role of the connector module in achieving robust cross-modal alignment for multimodal reasoning models. In addition, we introduce a unique indicator of reasoning capability, the entropy of critical reasoning tokens, which has proven highly effective for checkpoint selection during RL training. Skywork-R1V3 achieves state-of-the-art results of 76.0\% on MMMU. This performance matches entry-level human expert capabilities. Remarkably, our RL-powered post-training approach enables even the 38B parameter model to rival top closed-source VLMs. The implementation successfully transfers mathematical reasoning to other subject-related reasoning tasks. We also include an analysis of curriculum learning and reinforcement finetuning strategies, along with a broader discussion on multimodal reasoning. Skywork-R1V3 represents a significant leap in multimodal reasoning, showcasing RL as a powerful engine for advancing open-source VLM capabilities.

% \hfill
% \hfill

\begin{figure}[h!]
    \centering
    \includegraphics[width=1.0\linewidth]{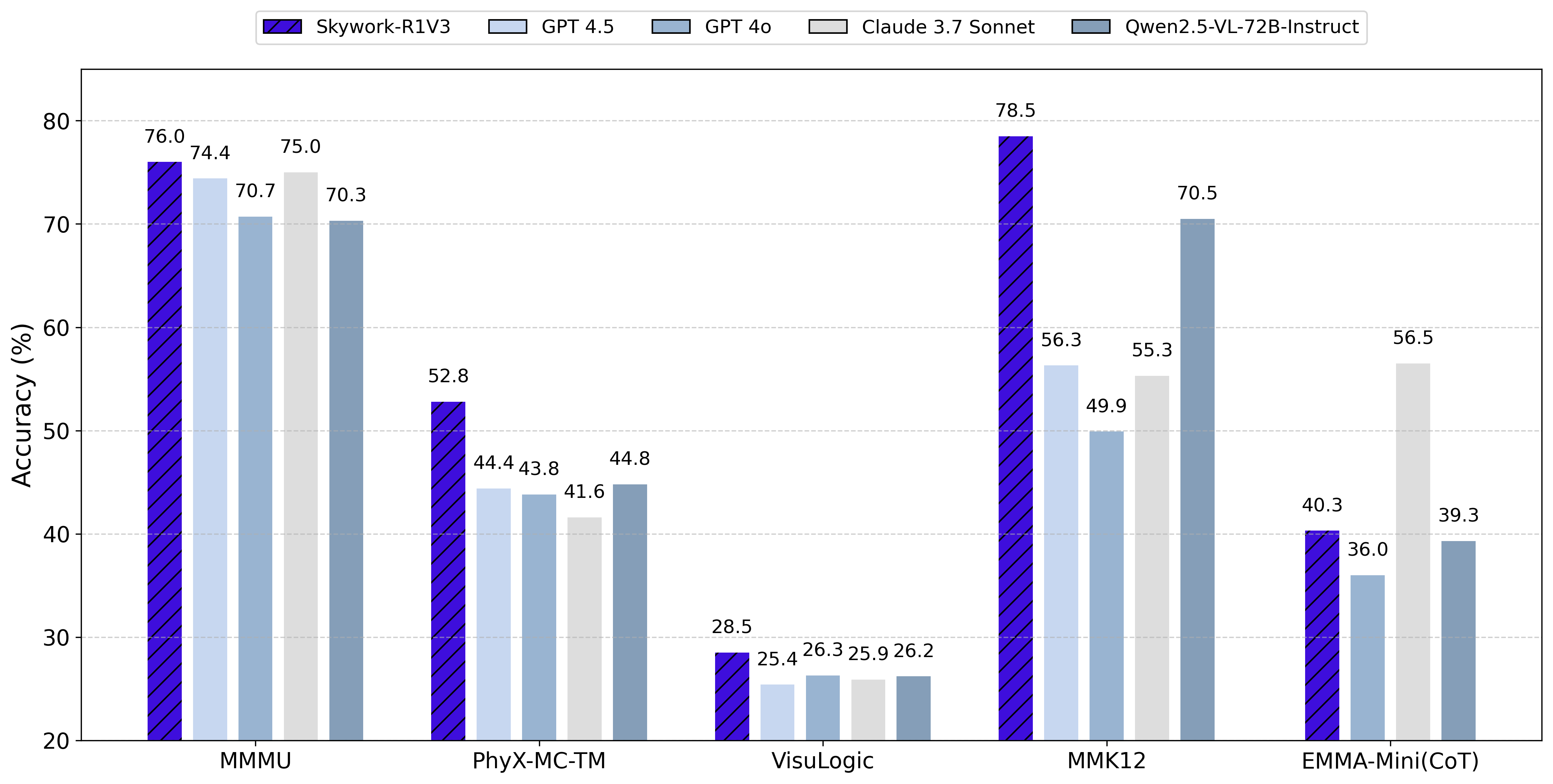}
    \caption{Benchmark performance of Skywork-R1V3.}
    \label{fig:enter-label}
\end{figure}
\vspace{-1em}
\end{abstract}

\newpage

\vspace{-4mm}
\section{Introduction}
Vision-language models (VLMs) have emerged as a foundational paradigm for general-purpose AI, enabling systems to perceive, reason, and act within open-ended virtual and physical environments. By aligning visual and textual modalities within a unified framework, VLMs have driven significant progress across diverse domains, including multimodal reasoning \citep{coreteam2025mimovltechnicalreport}, image editing \citep{zhou2025fireeditfinegrainedinstructionbasedimage}, GUI agents \citep{nguyen2024guiagentssurvey}, autonomous driving \citep{xu2024vlmadendtoendautonomousdriving}, and robotics \citep{wang2024vlmseerobotdo}. Beyond research, VLMs are increasingly powering real-world applications in education, healthcare, conversational AI, and wearable technologies \citep{Qwen2VL}. Despite this progress, current VLMs still lack human-level generality—particularly in tasks demanding robust spatial reasoning, precise object counting, imaginative visual synthesis, or complex interactive gameplay \citep{jian2025csvqachinesemultimodalbenchmark}. These limitations highlight persistent challenges in VLM development \citep{peng2025skyworkr1vpioneeringmultimodal}. Furthermore, the inherent heterogeneity of multimodal data complicates both training and inference, introducing bottlenecks in data pipeline design, distributed training optimization, and standardized evaluation.

This challenge stands in stark contrast to the situation for large language models (LLMs) \citep{touvron2023llamaopenefficientfoundation,deepseekai2024deepseekv3technicalreport}. While LLMs benefit from vast, high-quality textual corpora encoding broad human knowledge, VLMs suffer from scarce and uneven vision-language annotations—particularly for perceptually-grounded concepts such as texture, lighting, and spatial relations. \citep{zhu2025internvl3exploringadvancedtraining}. Consequently, the gap between closed-source and open-source models remains significantly wider for VLMs than for LLMs. In the LLM domain, strong open-source models like Deepseek R1\citep{deepseekai2025deepseekr1incentivizingreasoningcapability} provide high-quality Chain-of-Thought \citep{wei2022chain} reasoning, narrowing the capability gap. However, the VLM landscape faces a much steeper challenge in closing this divide. While proprietary models like Gemini Pro 2.5 \citep{gemini25}, OpenAI’s O-series \citep{jaech2024openai}, and Seed-VL\citep{guo2025seed15vltechnicalreport} demonstrate powerful visual reasoning across multiple domains, open-source alternatives such as QwenVL \citep{Qwen2VL}, InternVL \citep{zhu2025internvl3exploringadvancedtraining}, and LLaVA\citep{liu2023visualinstructiontuning} struggle to match their performance. Moreover, these open VLMs often lack comparable reasoning abilities and structured thinking patterns, further widening the gap.

In this report, we take a significant step forward in advancing open-source VLMs, particularly in complex visual reasoning tasks. Additionally, we provide comprehensive training details on post-training methodologies for VLMs based on open-source frameworks. Key points we address include:

\textbf{Cold Start Finetuning and RL can incentivize the main reasoning ability of the model.} 
We construct our cold-start dataset using early version Skywork R1V2 \citep{wang2025skyworkr1v2multimodalhybrid} and conduct SFT and RL experiments primarily with a instructed VLM. This strategy enables effective transfer of reasoning patterns from a language reasoning model to a vision-language model (Section \ref{sec:rl}).

\textbf{Critical token entropy indicates reasoning ability.}
In the RL training stage, we distinguish the models with genuine reasoning capabilities from those merely mimicking reasoning patterns through our novel metric, entropy of critical token. Specifically, we monitor the entropy values at critical reasoning initiation points. This metric strongly correlates with the actual reasoning performance in validation sets, providing an efficient method to identify high-quality checkpoints during RL training. (Section \ref{sec: critical_tokens}).

\textbf{Connector plays a central role in cross-modal alignment.}  
Prior work \citep{guo2025seed15vltechnicalreport} has primarily focused on achieving vision–text alignment during the pre-training phase and has demonstrated the effectiveness of the connector. In this study, we further identify the connector in VLMs as a critical component for maintaining cross-modal alignment during the reinforcement learning stage (Section \ref{sec: module activation of VLM}). Furthermore, we find that performing connector-only tuning after RL serves as an effective strategy to re-balance the model’s knowledge distribution without compromising its reasoning ability (Section \ref{mlp_after_rl}).

Beyond the core content, this report includes several additional topics: the application of curriculum learning to reinforcement learning (Section \ref{sec: curriculum_learning}), and reproductions of reinforcement learning tricks \citep{yu2025dapoopensourcellmreinforcement} such as Clip-Higher and Dynamic Sampling (Section \ref{sec: extra_rl}). We also explore the behavior of the visual reasoning model: the broader challenge of balancing memorization and generalization (Section \ref{sec: memory_or_general}), explore fast and slow thinking in inference (Section \ref{sec: fast_to_slow}), and investigate the effectiveness of thinking token budgets (Section \ref{sec: thining budget}) and hallucination in the reasoning transfer pipeline (Section \ref{sec: hallucination}). We hope these areas spark further investigation and engagement from the research community.

% \begin{center}
% \emph{
% Towards a Clean, Unified, and Aesthetically Pleasing arXiv Template
% }
% \end{center}

% \begin{figure}[h]
%     \centering
%     \includegraphics[width=\linewidth]{figures/figure1.png}
%     \vspace{-10mm}
%     \caption{We collaborated with Qwen and DeepSeek to complete this arxiv template.}
%     \label{fig:overall_figure}
% \end{figure}

\label{sec:intro}
\vspace{-1mm}

\section{Data Preparation}\label{sec:data}

\begin{figure}
    \centering
    \includegraphics[width=1\linewidth]{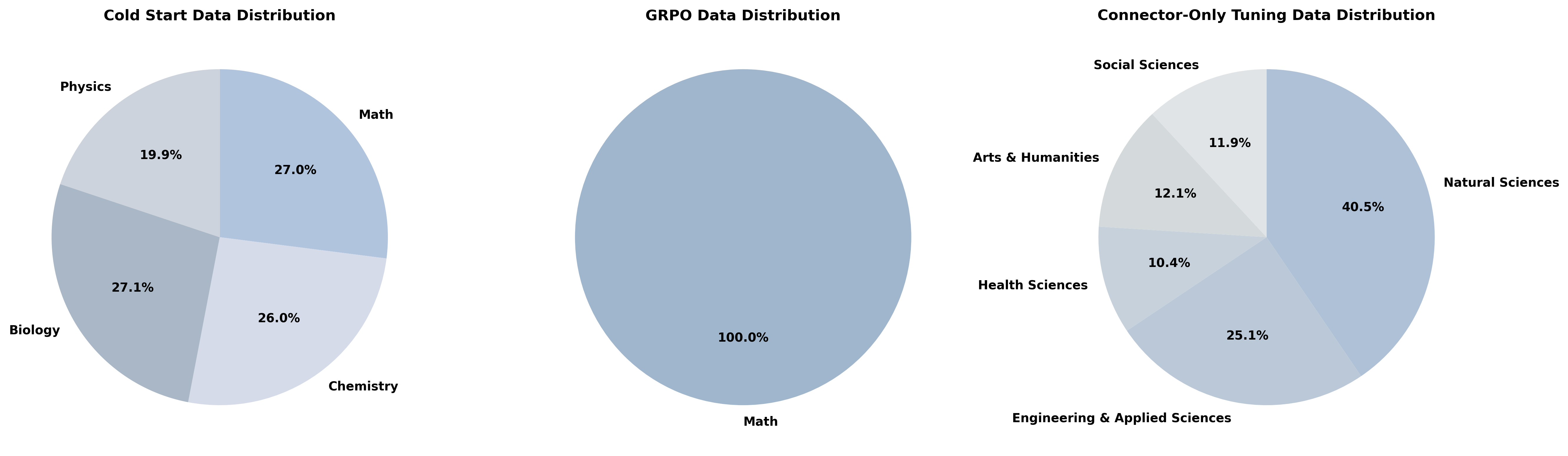}
    \caption{Data distribution across the three training stages.}
    \label{fig:data}
\end{figure}

\paragraph{LongCoT Data for Cold Start}  
We constructed a cold-start dataset consisting of approximately 20K instances collected from science-oriented practice questions of Chinese high-school difficulty, administered prior to 2024. The dataset covers four major disciplines: physics, chemistry, biology, and mathematics. It includes both multiple-choice diagnostic items and open-ended reasoning tasks.  

To create the corresponding reasoning processes, we leveraged the Skywork-R1V2 model to generate step-by-step thought processes for each question. We then applied a rejection sampling strategy to select only those instances where the model's final answer matched the ground-truth answer. This procedure resulted in a filtered dataset containing approximately 12k high-quality examples with reasoning chains.

\paragraph{Data for RL}  
For the RL stage, we curated 15K high-quality multimodal \textit{\textbf{math}} data with K12-level difficulty. The dataset consists entirely of multiple-choice and fill-in-the-blank items, where each instance is represented as a question $q$ paired with its corresponding answer $a$, and no explicit reasoning steps are included.

\paragraph{Data for Connector-only Tuning}
To ensure domain diversity, we selected 10K examples from 20 distinct domains, including:

\begin{itemize}
\item \textbf{Natural Sciences:} Mathematics, Physics, Chemistry, Biology
\item \textbf{Engineering \& Applied Sciences:} Engineering, Agriculture, Environmental Science, Computer Science
\item \textbf{Health Sciences:} Medicine, Pharmacy
\item \textbf{Arts \& Humanities:} Music, Literature, Philosophy, History
\item \textbf{Social Sciences:} Economics, Psychology, Political Science, Sociology, Education, Law
\end{itemize}

The data distribution across the three training stages is illustrated in the Figure \ref{fig:data}.

% \subsection{Reinforcement Learning witb Verifiable Rewards}

\section{Post-Training Recipes}\label{sec:rl}
Our post-training recipes mainly consist of reward function design, cold start finetuning, reinforcement finetuning and connector-only finetuning. We will discuss details in this section.

\subsection{Reward Function Design}
Our reward function typically consists of a format reward and an accuracy reward. The format reward is defined as:
\begin{equation}
R_{\text{format}}(y) = \mathbb{1} (c = 1)
\end{equation}

where $ c = 1 $ indicates that the generated response adheres to the specified chat template, which follows the structure ``\texttt{<think>...<\/think>...}''.

The accuracy reward is the primary objective of our Reinforcement Learning (RL) process, which we aim to maximize:
\begin{equation}
R_{\text{accuracy}}(\hat{a}, a) = \begin{cases}
1, & \text{if } R_{\theta}(\hat{a}, a) = 1 \text{ or } R_{\text{rule}}(\hat{a}, a) = 1 \\
0, & \text{else}
\end{cases}
\end{equation}

Here, $R_{\theta}(\hat{a}, a)$ likely represents an accuracy score derived from a  language model-based verifier (eg. Qwen2.5-Instruct 32B), and $R_{\text{rule}}(\hat{a}, a)$ represents an accuracy score based on predefined rules.

Therefore, our final reward is defined as:
\begin{equation}
R = \epsilon R_{\text{accuracy}} + (1 - \epsilon)R_{\text{format}}
\end{equation}

Empirically, we set $\epsilon$ to 0.8 in this report, emphasizing accuracy as the primary reward component.

\subsection{Cold Start Finetuning}
To initiate training, we collected thousands of cold-start samples from an early internal version of Skywork-R1V2. While demonstrating strong reasoning capabilities, this model suffered from issues like poor readability and language mixing. To enhance reasoning clarity and enable open sharing, we employed the Skywork-VL-Reward \citep{wang2025skyworkvlrewardeffectivereward} alongside GPT-4o to filter rambling and overly lengthy samples, resulting in a refined cold-start dataset.

Significantly, Skywork-R1V2 is created by directly stitching InternViT-6B-448px-V2.5 \citep{zhu2025internvl3exploringadvancedtraining} and QwQ-32B \citep{qwq32b}. Its visual reasoning capabilities originate from textual reasoning, illustrating a successful transfer of reasoning skills and the potential for cross-modal reasoning inheritance.

Building on this, we integrate R1V2's reasoning style and validated visual reasoning traces into the SFT stage using InternVL-38B \citep{zhu2025internvl3exploringadvancedtraining} as the base model. Following the cold-start phase, the model naturally adopted the "think before answering" (R1-style) approach—an inherent capability present from the LLM's pretraining stage. By preserving the valid visual reasoning distribution identified in R1V2, we further enhance the model's reasoning proficiency.

\subsection{Reinforcement Finetuning}

The RL objective is to learn a reasoning policy $\pi_\theta$ that maximizes the expected reward across prompts sampled from distribution $\mathcal{D}$, i.e.:
\begin{equation}
    \max_\pi J(\pi) := \mathbb{E}_{x \sim \mathcal{D}, y \sim \pi(\cdot | x)} [r(x, y)].
\end{equation}

\paragraph{PPO}
Proximal Policy Optimization (PPO)~\citep{schulman2017proximalpolicyoptimizationalgorithms} is a state-of-the-art policy gradient algorithm that ensures training stability by constraining the size of policy updates. It optimizes a clipped surrogate objective function, which is a lower bound on the policy performance improvement. The core objective for PPO is given by:
\begin{equation}
    \mathcal{L}^{\text{PPO}}(\theta) = \hat{\mathbb{E}}_t \left[
        \min \left(
            \rho^t(\theta) \hat{A}^t, \,
            \mathrm{clip}(\rho^t(\theta), 1 - \epsilon, 1 + \epsilon) \hat{A}^t
        \right)
    \right].
\end{equation}
Here, $\rho^t(\theta) = \frac{\pi_\theta(a^t | s^t)}{\pi_\text{ref}(a^t | s^t)}$ is the importance sampling ratio between the new policy $\pi_\theta$ and the old policy $\pi_k$ used for data collection. $\hat{A}^t$ is an estimator of the advantage function at timestep $t$, and $\epsilon$ is a hyperparameter that defines the clipping range.

\paragraph{GRPO}
Group-normalized Reward Policy Optimization (GRPO) \citep{shao2024deepseekmathpushinglimitsmathematical} is a variant of PPO tailored for learning from sparse, binary rewards ($r(x, y) \in \{0, 1\}$), often provided by a verifier. The key innovation of GRPO lies in its advantage estimation method.

Instead of using the raw binary reward, GRPO first samples $M$ independent and identically distributed candidate traces $\{y_{i,1}, ..., y_{i,M}\}$ from the current policy $\pi_k$ for each input $x_i$. The token-level advantage $A^t_{ij}$ for each trace $y_{ij}$ is then estimated by normalizing its final reward within the sampled group:
\begin{equation}
    A^t_{ij} := \frac{r(x_i, y_{ij}) - \text{mean}(r(x_i, y_{i,1}), ..., r(x_i, y_{i,M}))}{\text{std}(r(x_i, y_{i,1}), ..., r(x_i, y_{i,M}))},
\end{equation}
 This group normalization converts the sparse binary signal into a dense, continuous advantage estimate, providing a richer learning signal for the policy.

The final GRPO policy loss incorporates this group-normalized advantage into the PPO objective, adding a KL-divergence penalty that regularizes the policy $\pi_\theta$ towards a reference policy $\pi_{\text{ref}}$:
\begin{equation}
\begin{split}
\mathcal{L}^{\text{GRPO}}(\theta) = -\hat{\mathbb{E}}_{x_i, \{y_{ij}\}}
&\left[
\frac{1}{M} \sum_{j=1}^M \frac{1}{|y_{ij}|} \sum_{t=0}^{|y_{ij}|-1}
\left( \min \left( \rho^t_{ij}(\theta) A^t_{ij}, \, \mathrm{clip}\left(\rho^t_{ij}(\theta), 1 - \epsilon, 1 + \epsilon\right) A^t_{ij} \right) \right. \right. \\
&\left. \left. \qquad\qquad\qquad - \beta D^t_{ij}(\theta) \right)
\right].
\end{split}
\end{equation}
In this equation, $\rho^t_{ij}(\theta) = \frac{\pi_\theta(a^t_{ij} | s^t_{ij})}{\pi_\text{ref}(a^t_{ij} | s^t_{ij})}$ is the importance sampling ratio, $\epsilon$ is inherited from PPO clipping parameter, $\beta$ controls the strength of the KL regularization, and $D^t_{ij}(\theta) = \log \frac{\pi_{\theta}(a^t_{ij} | s^t_{ij})}{\pi_\text{ref}(a^t_{ij} | s^t_{ij})} $ is the KL-divergence penalty.

\paragraph{RL Implementation Details}
We use the VERL framework \citep{sheng2024hybridflow} for RL training, progressively increasing the context and output lengths from 4,096 to 8,192. Each episode samples 1,024 rollouts, with 32 verifier-rewarded samples per prompt. Training employs a mini-batch size of 32, 1 gradient step per episode (on-policy finetuning for performance maximization), and a PPO clipping range of 0.2. No KL divergence penalty is applied during long CoT training .

The actor's learning rate is set to 1e-5, differing from costumed GRPO configurations (discussed in Section \ref{sec: learning rate strategy}). Besides, we keep all modules trainable during RL—Section \ref{sec: module activation of VLM} analyzes this choice.

\subsection{Connector-Only Tuning}

RL training encourages the model to explore and optimize for reward-related behaviors such as correctness and format coherence, which has proven to be a highly effective approach for enhancing the reasoning abilities of VLM.  However, since the training data during the reinforcement learning stage were primarily composed of mathematical problems, the model’s knowledge acquisition exhibited a certain bias: while its reasoning ability improved, its knowledge base became skewed towards the mathematics domain.

To rebalance the knowledge base and enhance the model’s cross-disciplinary reasoning capabilities, we introduced an additional finetuning step targeting the cross-modal connector after the reinforcement learning stage. Specifically, we constructed a high-quality, multi-domain, multimodal dataset of 10K samples (see Section \ref{sec:data}) to retrain the connector in a targeted manner, optimizing its ability to integrate knowledge across diverse domains.

This step effectively balanced the model’s knowledge distribution, significantly improving its perception and understanding in non-mathematical fields such as humanities, medicine, and the arts, while maintaining its original strengths in reasoning. As a result, the model achieved stronger generalization and reasoning performance across a wide range of disciplines.

\paragraph{Implementation Details}
Training was conducted for two epochs using the 10K multi-domain multimodal dataset, with a global batch size of 64 and a learning rate of 1e-5, scheduled using a cosine decay with a 0.03 warmup ratio. The maximum sequence length was set to 16,384 tokens. Early stopping is applied based on validation loss to avoid overfitting.

% RL often induces overthinking in model responses. In particular, models tend to produce overly verbose answers, characterized by unnecessary elaboration, redundant reasoning chains, or speculative content, even when the question requires a direct or succinct answer.

% To mitigate these issues, we adopt a lightweight post-alignment strategy. The goal is to rebalance the model's behavior across domains and response styles by leveraging a curated, balanced dataset representing diverse disciplines. This strategy serves as a behavioral regularizer—aligning the model to desired traits such as conciseness and cross-domain consistency—while preserving the core reasoning competencies acquired through RL.

% \subsection{Dataset Construction}

% To construct high-quality instruction–response pairs for each domain, we adopted a distillation pipeline based on \jp{Gemini/self-distillation}. Candidate responses were generated for each prompt, and a reject sampling strategy was employed to retain only those outputs that met predefined correctness and relevance criteria. This process ensured that the resulting dataset consisted of accurate and well-aligned responses.

% To further enhance response quality, we applied a filtering procedure to eliminate samples exhibiting excessive verbosity. Specifically, responses characterized by unnecessarily long reasoning chains, redundant elaborations, or speculative content were discarded in favor of concise and focused outputs.

% 

% \section{Strong-to-Weak Reasoning Transferring} 

% [38B->8B  8B->training from scratch]

\begin{table}[ht]
\centering
\small
\caption{Comparison of models on visual-language benchmarks}
\begin{threeparttable}
\begin{tabular}{lccccc|cc}
\toprule
\textbf{Benchmark} & \makecell{\textbf{Metrics}\\\textbf{(Avg@5)}}& \makecell{\textbf{Skywork-}\\\textbf{R1V3-38B}} & \makecell{\textbf{QVQ-72B}\\\textbf{Preview}} & \textbf{Internvl3-78B} & \makecell{\textbf{Qwen2.5}\\\textbf{VL 72B}} & \makecell{\textbf{Claude}\\\textbf{3.7 Sonnet}} & \makecell{\textbf{GPT-4o}} \\
\midrule
\rowcolor{teal!15} \multicolumn{8}{l}{\textit{\textbf{General}}} \\
MMMU\tnote{val} & Acc. & \textbf{76.0} & 70.3 & 72.2 & 70.3 & 75.0 & 70.7 \\
EMMA\tnote{mini-cot} & Acc. & 40.3 & 32.0 & 38.3 & 39.3 & \textbf{56.5} & 36.0 \\
MMMU-pro & Acc. & \textbf{55.4} & 46.9* & 48.6 & 51.1 & 50.0 & 54.5 \\
MMK12 & Acc. & \textbf{78.5} & 62.7* & 67.4* & 70.5* & 55.3 & 49.9 \\
MMstar & Acc. & 70.6 & 60.8 & \textbf{72.5} & 70.8 & 68.8 & 65.1 \\
MMBench-en-1.1 & Acc. & 85.7 & 72.6* & 87.7 & \textbf{88.0} & 82.0 & 84.3 \\
HallusionBench & Acc. & \textbf{61.3} & 55.3* & 59.1 & 55.2 & 58.3 & 56.2 \\
\rowcolor{teal!15} \multicolumn{8}{l}{\textit{\textbf{Mathematics}}} \\
MathVista\tnote{mini} & Acc. & \textbf{77.1} & 71.4 & 72.2 & 74.8 & 66.8 & 62.9 \\
MathVerse\tnote{vision-only} & Acc. & \textbf{59.6} & 45.1 & 51.0 & 57.6 & 49.9* & 49.9 \\
MathVision & Acc. & 52.6 & 35.9 & 43.1 & 38.1 & \textbf{58.6} & 31.2 \\
WeMath\tnote{strict} & Acc. & \textbf{56.5} & 37.7 & 46.1 & 50.6 & 48.9* & 50.6 \\
\rowcolor{teal!15} \multicolumn{8}{l}{\textit{\textbf{Logic}}} \\
Visulogic & Acc. & \textbf{28.5} & 23.5* & 27.7 & 26.2 & 25.9 & 26.3 \\
LogicVista & Acc. & 59.7 & 53.8 & 55.9 & 57.1 & 60.6* & \textbf{64.4} \\
MME-reasoning & Acc. & \textbf{42.8} & 35.2 & 32.1 & 34.1 & 34.1 & 30.2 \\
\rowcolor{teal!15} \multicolumn{8}{l}{\textit{\textbf{Physics}}} \\
PhyX\tnote{mc-text-minimal} & Acc. & \textbf{52.8} & 35.2* & 40.5 & 44.8 & 41.6 & 43.8 \\
SeePhys & Acc. & 31.5 & 22.5 & 19.0* & 24.2 & \textbf{34.6} & 21.9 \\
\hline

\end{tabular}
\begin{tablenotes}
\footnotesize
\item[*] indicates results from our evaluation framework.
\end{tablenotes}
\end{threeparttable}
\end{table}
\section{Evaluation}
To comprehensively evaluate the reasoning capabilities of Skywork-R1V3, we assess its performance on a diverse suite of publicly available multimodal benchmarks. These benchmarks encompass both general-purpose understanding tasks and advanced multimodal reasoning tasks, thereby providing a broad and rigorous evaluation setting. Furthermore, we position Skywork-R1V3-38B relative to several mainstream multimodal models on representative public benchmarks to offer a more holistic perspective on its overall performance.

\paragraph{Evaluation Settings} 
\begin{wrapfigure}{r}{0.5\textwidth}  % r: 右对齐，0.4\textwidth 控制图片宽度
    \centering
    \includegraphics[width=\linewidth]{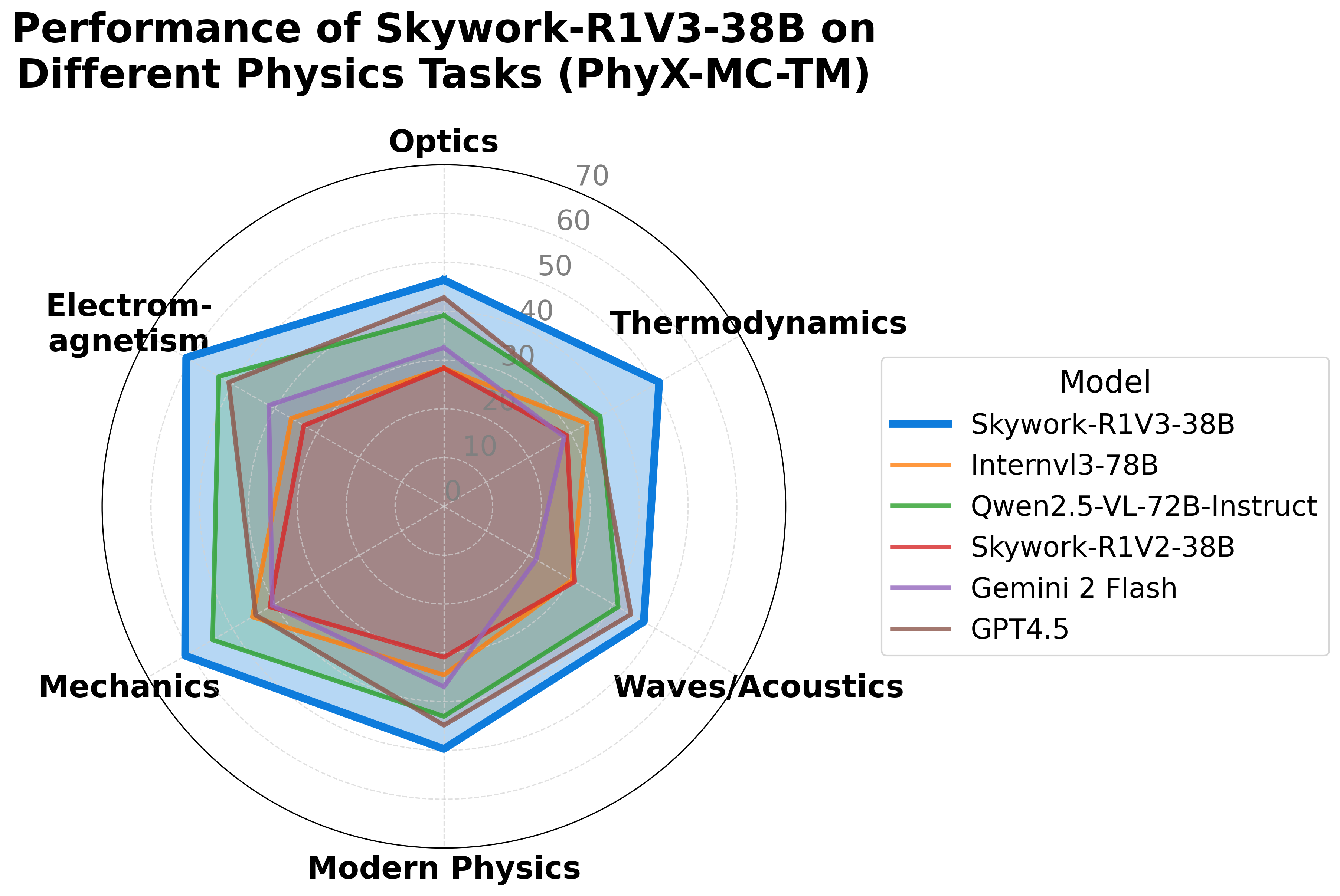}
    \caption{The Performance of Skywork-R1V3-38B on PhyX-MC-Text-Minimal}
    \label{fig:phy}
\end{wrapfigure}
For evaluation, we use a generation token budget of 16,384 and a decoding temperature of 1.0. Results are averaged over five runs. We adopt VLMEvalKit \citep{duan2024vlmevalkit} as the primary evaluation framework, with targeted modifications (i.e. add rule-based check for MMMU \citep{yue2024mmmu}) to better support long chain-of-thought reasoning models. Additionally, we refine task-specific evaluation protocols to improve consistency and reliability across benchmarks. To promote open research and ensure reproducibility, we plan to release the complete evaluation framework along with all associated prompt sets. Our evaluations compare against a diverse set of representative models, including open-source models (InternVL3-78B~\citep{zhu2025internvl3exploringadvancedtraining}, Qwen2.5-VL-72B~\citep{Qwen2VL} and QVQ-72B Preview~\citep{qwen2024qvq}) and leading proprietary models (Claude 3.7 Sonnet~\citep{Claude3.7}, GPT-4o~\citep{openai2024gpt4o}), to ensure a comprehensive and rigorous benchmarking.\par

\begin{figure}[!ht]
    \centering
    \includegraphics[width=0.7\linewidth]{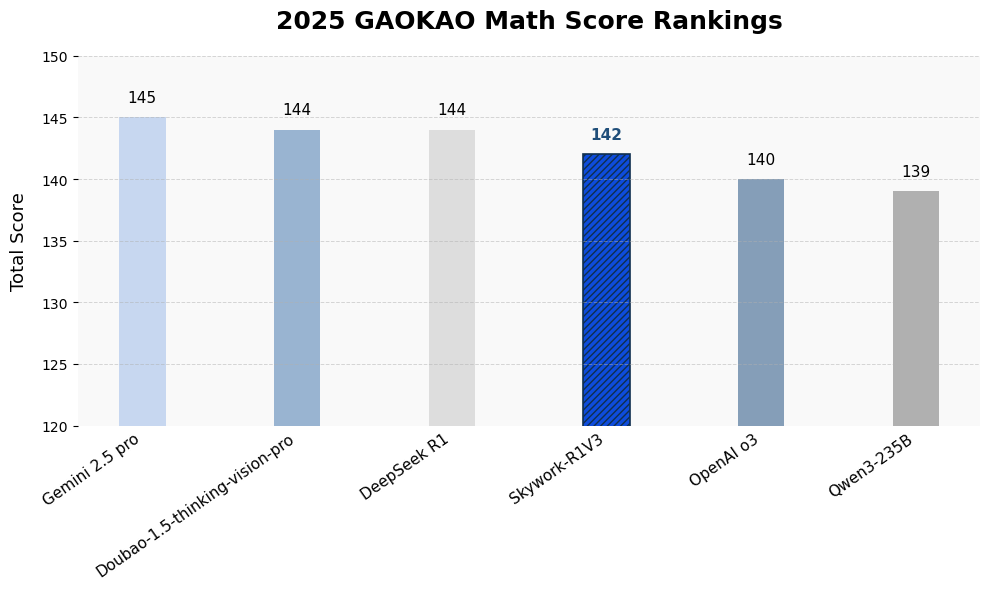}
    \caption{Model Rankings on 2025 GAOKAO Math}
    \label{fig:gaokao}
\end{figure}

\paragraph{General Benchmarks} 
On general vision-language benchmarks, Skywork-R1V3-38B demonstrates strong and consistent performance, achieving state-of-the-art among open-source models on MMMU \citep{yue2024mmmu} (76.0\%), MMMU-pro \citep{yue2025mmmuprorobustmultidisciplinemultimodal} (55.4\%), and MMK12~\citep{meng2025mm} (78.5\%).
On EMMA~\citep{hao2025can}, Skywork-R1V3-38B  surpasses larger open-source models such as InternVL3-78B and Qwen2.5-VL-72B.
On MMStar~\citep{chen2024we}, Skywork-R1V3-38B reaches 70.6\%, which is comparable to the best open-source model (InternVL3-78B at 72.5\%). Similarly, on MMBench-en-1.1~\citep{liu2024mmbench}, Skywork-R1V3-38B scores 85.7\%, ranking among the top performers and closely matching the larger InternVL3-78B (87.7\%) and Qwen2.5-VL-72B (88.0\%). In HallusionBench~\citep{guan2024hallusionbenchadvanceddiagnosticsuite} (which evaluates multimodal hallucination robustness), Skywork-R1V3-38B achieves 61.3\%, outperforming other advanced models and demonstrating solid reliability in this challenging scenario.
\paragraph{Mathematics Benchmarks}  In the mathematics reasoning benchmarks, Skywork-R1V3-38B exhibits robust multi-modal problem-solving abilities. For instance, it achieves 77.1\% on MathVista~\citep{lu2023mathvista}, 59.6\% on MathVerse \citep{zhang2024mathversedoesmultimodalllm} and 56.5\% on WeMath~\citep{qiao2024wemathdoeslargemultimodal} (strict), consistently outperforming all larger open-source models (72B/78B) on these tasks. Skywork-R1V3-38B also surpasses Claude
3.7 Sonnet and GPT-4o on most of these math benchmarks, highlighting the strength of its advanced reasoning capabilities in complex mathematics problems. Skywork-R1V3-38B achieves a score comparable to top human performance on the 2025 GAOKAO math exam (see Figure \ref{fig:gaokao}), demonstrating exceptional reasoning and generalization abilities on out-of-distribution (OOD) problems from the Chinese College Entrance Exam. Notably, while our training data primarily consists of English content, R1V3 attains state-of-the-art (SOTA) performance among 38B-scale models on this benchmark. Evaluated through a rigorous human-LLM mixed-judging approach, Skywork-R1V3 scored 142 out of 150, surpassing several advanced models in the field.

\paragraph{Logic Benchmarks} Our Skywork-R1V3-38B continues to perform competitively on logical reasoning tasks, achieving 28.5\% on VisuLogic~\citep{xu2025visulogicbenchmarkevaluatingvisual} and narrowly outperforming the best open competitor (InternVL3-78B at 27.7\%). On the  LogicVista~\citep{xiao2024logicvistamultimodalllmlogical} benchmark, Skywork-R1V3-38B reaches 59.7\%, approaching the performance of GPT-4o (64.4\%) while surpassing all other open-source models. Furthermore, Skywork-R1V3-38B achieves 42.8\% on the MME-Reasoning~\citep{yuan2025mmereasoningcomprehensivebenchmarklogical} benchmark, which exceeds the larger open-source model (QVQ-72B Preview, 35.2\%) by over 7 points and outperforms GPT-4o (30.2\%) by more than 12 points.

\paragraph{Physics Benchmarks} 
% 在需要插入图片的位置添加以下代码

On physics-related benchmarks, Skywork-R1V3-38B demonstrates strong generalization. It achieves 52.8\% on PhyX~\citep{shen2025phyxdoesmodelwits}, substantially outperforming competing models (detailed performance is shown in Figure \ref{fig:phy}). On SeePhy~\citep{xiang2025seephysdoesseeinghelp}, Skywork-R1V3-38B scores 31.5\%, which is just below Claude 3.7  Sonnet’s 34.6\% but higher than other open-source models. These results indicate that our Skywork-R1V3-38B can effectively handle complex physics problems, outperforming much larger open-source models and even approaching the advanced proprietary model’s performance.

% \input{tables/main_table}

% \usepackage{wrapfig}  % 在导言区添加

% \begin{wrapfigure}{r}{0.5\linewidth}  % r: 右对齐，0.5为宽度
%     \centering
%     \includegraphics[width=\linewidth]{figures/phyX.png}
%     \caption{Enter Caption}
%     \label{fig:phy}
% \end{wrapfigure}

%\subsection{The Effectiveness of Thinking Budget}

\vspace{-2mm}
\section{Empirical Analysis on Reinforcement Learning}
This section details our empirical analysis of Reinforcement Learning (RL) applied to Skywork-R1V3. RL demonstrates great potential for enhancing model capabilities. Our findings aim to stimulate further research into equipping VLMs with advanced reasoning capabilities via RL.  For this study, we utilized the GRPO algorithm.

We highlight that direct transfer of RL techniques from text-only Large Language Models (LLMs) to VLMs is insufficient due to the distinct challenges posed by visual modality integration. This necessitates the development of VLM-specific RL recipes. Ultimately, this section underscores the critical need for continued exploration to optimize RL for Visual Large Language Models (VLMs) and unlock their full potential in visual language understanding and reasoning.

\subsection{Critical Token Entropy Indicates Reasoning Ability}
\label{sec: critical_tokens}
% \wshen{detailed analaysis on the length  and entropy variance}

\begin{figure}[!htbp]
    \centering
    \begin{subfigure}[b]{0.48\textwidth}
        \centering
        \includegraphics[width=\linewidth]{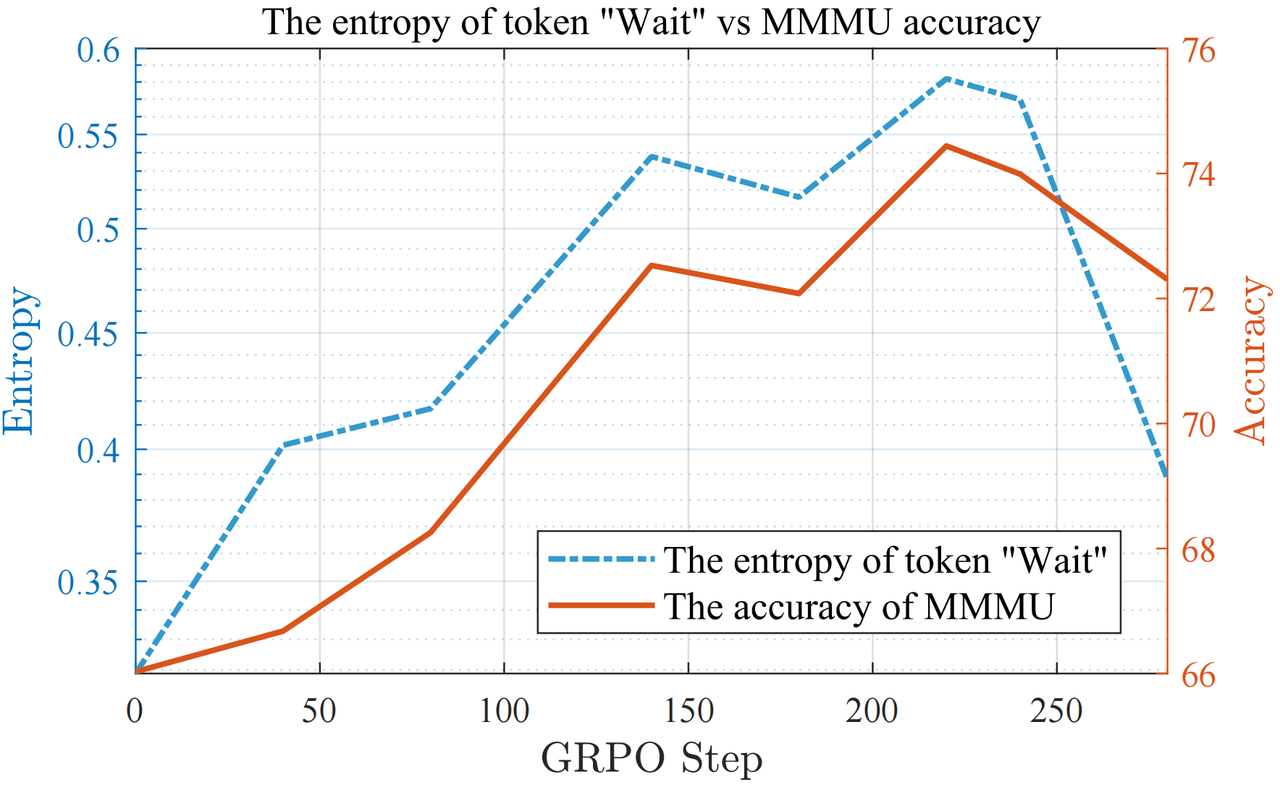}
        % \caption{The entropy of token ''}
        \label{fig:sub1}
    \end{subfigure}
    \hfill
    \begin{subfigure}[b]{0.48\textwidth}
        \centering
        \includegraphics[width=\linewidth]{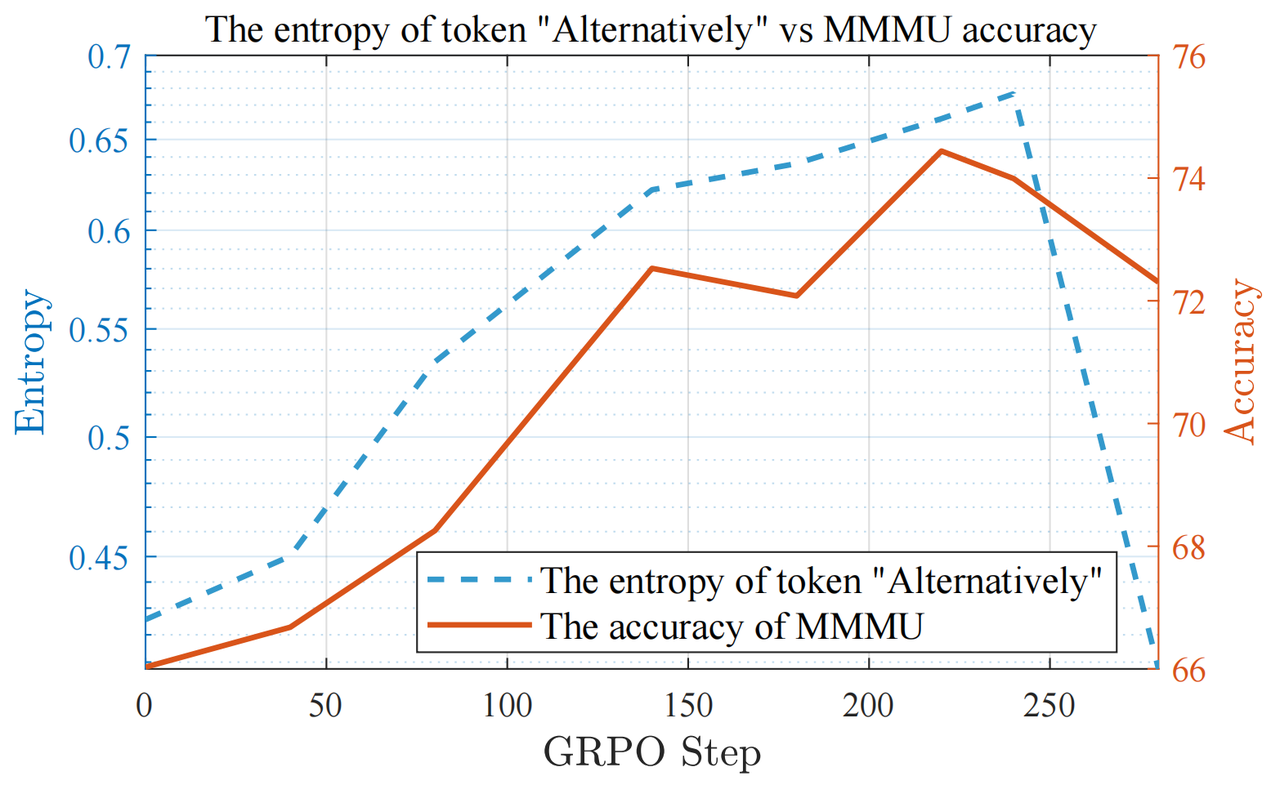}
        % \caption{Second Subfigure Caption}
        \label{fig:sub2}
    \end{subfigure}
    \caption{The entropy of critical token vs. MMMU accuracy}
    \label{fig:crucial token_entropy}
\end{figure}
% It is widely recognized that reinforcement finetuning suffers from inherent instability, which can manifest as model collapse or reward hacking. Specifically, Reinforcement Learning from Verified Reward (RLVR) is often implicated in model collapse due to the issue of advantage diminish, while Reinforcement Learning from Human Feedback (RLHF) is prone to reward hacking, a consequence of the unrobustness and potential ambiguity of human-assigned rewards, especially for out-of-distribution (OOD) samples.

% The core distinction between these methodologies lies in their respective reward functions. In practice, ensuring training saturation and training stability is paramount for achieving sustained and robust RL performance. To this end, we incorporate length variance and entropy variance as key performance indicators for our RL framework. The entropy variance is significant to supervise the anomaly beharviour of the training. If the entropy variacne very large as depicted in Figure \ref{}. The model can quickly collapse after the point which huge entropy variance, with the gradient norm and KL is extremely huge unstable,

During the reinforcement learning training process, we observed that although models fine-tuned with cold start supervised learning can generate responses that appear to follow a reasoning style and achieve high reward values on in-distribution training and validation sets, in most cases, this merely reflects the model repeating existing patterns rather than truly activating generalizable reasoning capabilities. To address this, Skywork-R1V3 introduces a unique metric, which utilizes the entropy of critical tokens to validation. Specifically, we focus on the output entropy values at critical positions where the model begins reasoning (such as when generating tokens like "Wait..." or "Alternatively..." to enter the reasoning process). Models with genuine reasoning capabilities typically exhibit high uncertainty at these positions, demonstrating characteristics of divergent thinking. In contrast, models that merely mimic reasoning style and tone usually produce low-entropy, deterministic content at these key points.  

Based on this insight, we can efficiently identify model checkpoints that have truly learned to reason during reinforcement training while filtering out those that simply "follow the script." As depicted in Figure \ref{fig:crucial token_entropy}, the level of entropy at these critical points shows a strong correlation with the model’s actual reasoning performance on the validation set. This mechanism provides a novel and efficient method for model selection during reinforcement training, ensuring that the final chosen model weights have indeed acquired generalizable reasoning abilities.

% \subsection{Entropy Control}

\subsection{The Connector Module Activation is Vital in RL}
\label{sec: module activation of VLM}
\begin{figure}[!h]
    \centering
    \includegraphics[width=1\linewidth]{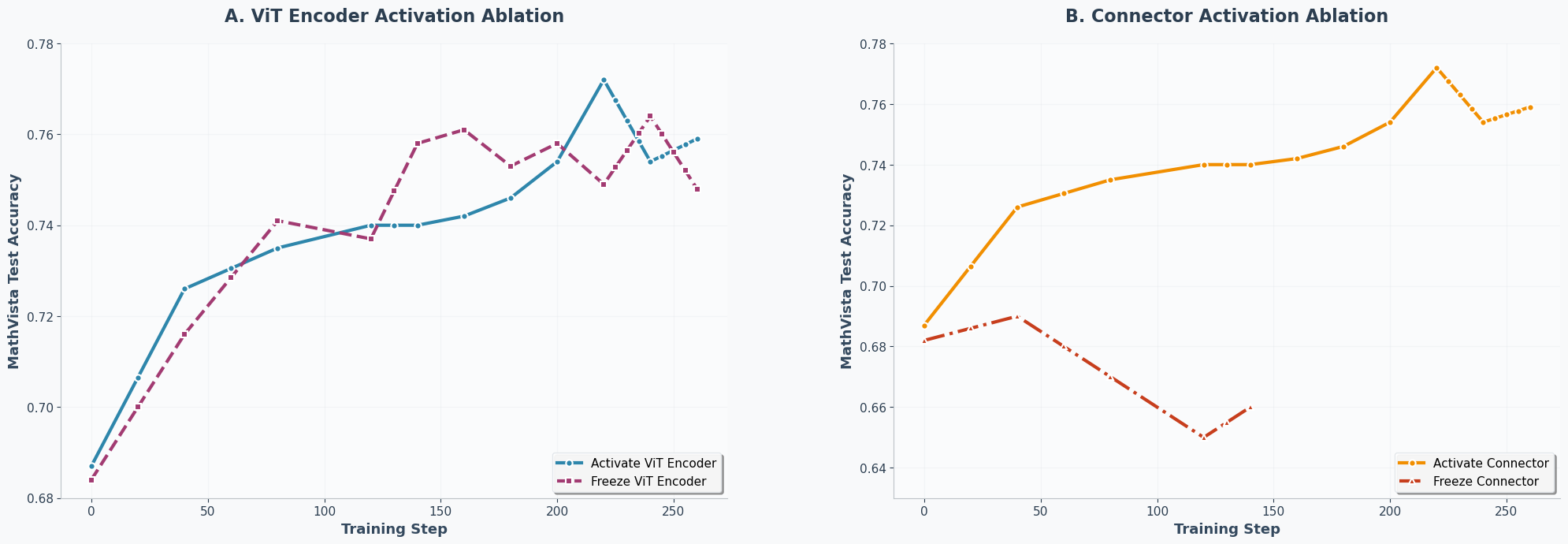}
    \caption{Ablation Studies of Module Activation Impact on MathVista Performance}
    \label{fig:connector_activation}
\end{figure}

Here, we investigate primarily the activation of the Skyword-R1V3 module in RL. The core difference between multimodal reasoning reinforcement learning and text-based reasoning reinforcement learning lies in the connector and the visual encoder. During reinforcement learning, we observed that whether the visual encoder is activated (with trainable parameters) has a significant impact on the final performance of R1V3. 

\paragraph{Visual Encoder Activate vs. Freeze} With the connector remaining trainable, whether or not the visual encoder parameters are updated, the reward curve remains largely consistent (depicted in Figure \ref{fig:connector_activation}A). This indicates that freezing the visual encoder does not disrupt the normal progression of the training process. However, allowing the visual encoder to participate in the training continues to bring additional performance gains. Even if the visual encoder has been well-pretrained, finetuning it during the reinforcement learning phase for specific cross-modal reasoning tasks remains valuable, as it helps the model capture more task-relevant representational details.

\paragraph{Connector Activate vs. Freeze} We found that the trainability of the connector is an absolute prerequisite for stable model learning. Once the connector is frozen or removed during the reinforcement learning phase, the model quickly experiences training failure: the reward curve drops sharply, the unstable gradient norm, and the model output degenerates into meaningless repetitive text, completely losing its reasoning ability (as shown in Figure \ref{fig:connector_activation}B).

% Todo [Figure 1: Comparison of reward curves with different module activation states during RL training - including four curves: visual encoder activation/freeze, connector activation/freeze, clearly showing the collapse caused by freezing the connector]

This experiment demonstrates that the connector, as the core bridge integrating visual and language modalities, plays a fundamental role in cross-modal reasoning optimization. It ensures that the representations of the two heterogeneous modalities can be effectively guided, interacted with, and aligned during training, collectively pointing to the latent shared reality in the Platonic Representation Hypothesis \citep{huh2024platonicrepresentationhypothesis}. In contrast, the activation of the visual encoder acts as a supplementary optimization on this stable bridge, enhancing the model's representation accuracy for specific task details, but its role is not as foundational and critical as that of the connector.

\subsection{The Distribution Shift in Curriculum Learning Hinder Generalization}
\label{sec: curriculum_learning}
\begin{figure}[!h]
 
    \begin{subfigure}[b]{0.48\linewidth} % b for bottom alignment
        \includegraphics[width=\linewidth]{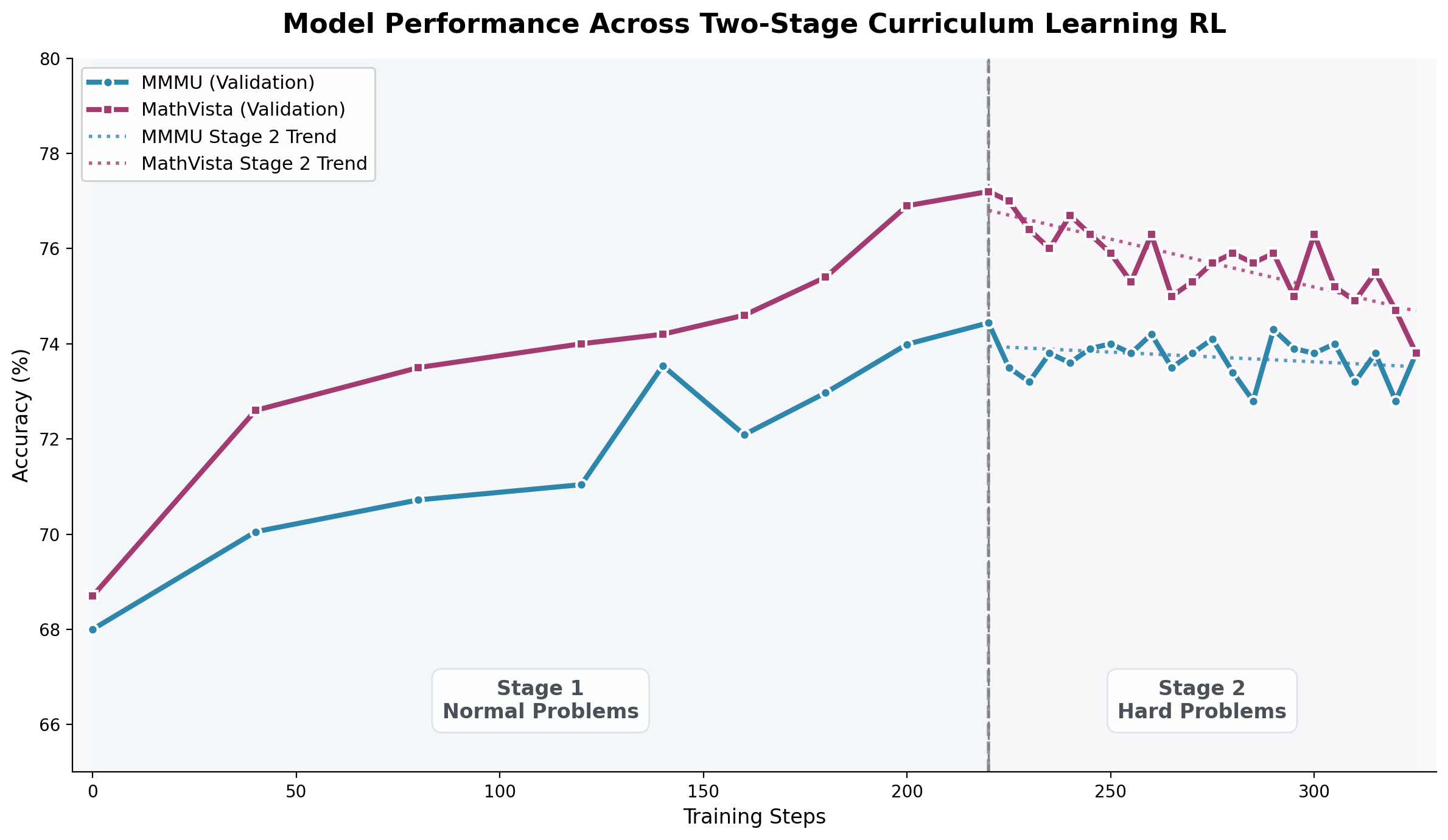}
        \caption{Overall Curriculum Learning Progress}
        \label{fig:curriculum_learning}
    \end{subfigure}
    \hfill % This command distributes horizontal space evenly between subfigures
       \centering
    \begin{subfigure}[b]{0.48\linewidth} % b for bottom alignment, 0.48\linewidth to leave space for separation
        \includegraphics[width=\linewidth]{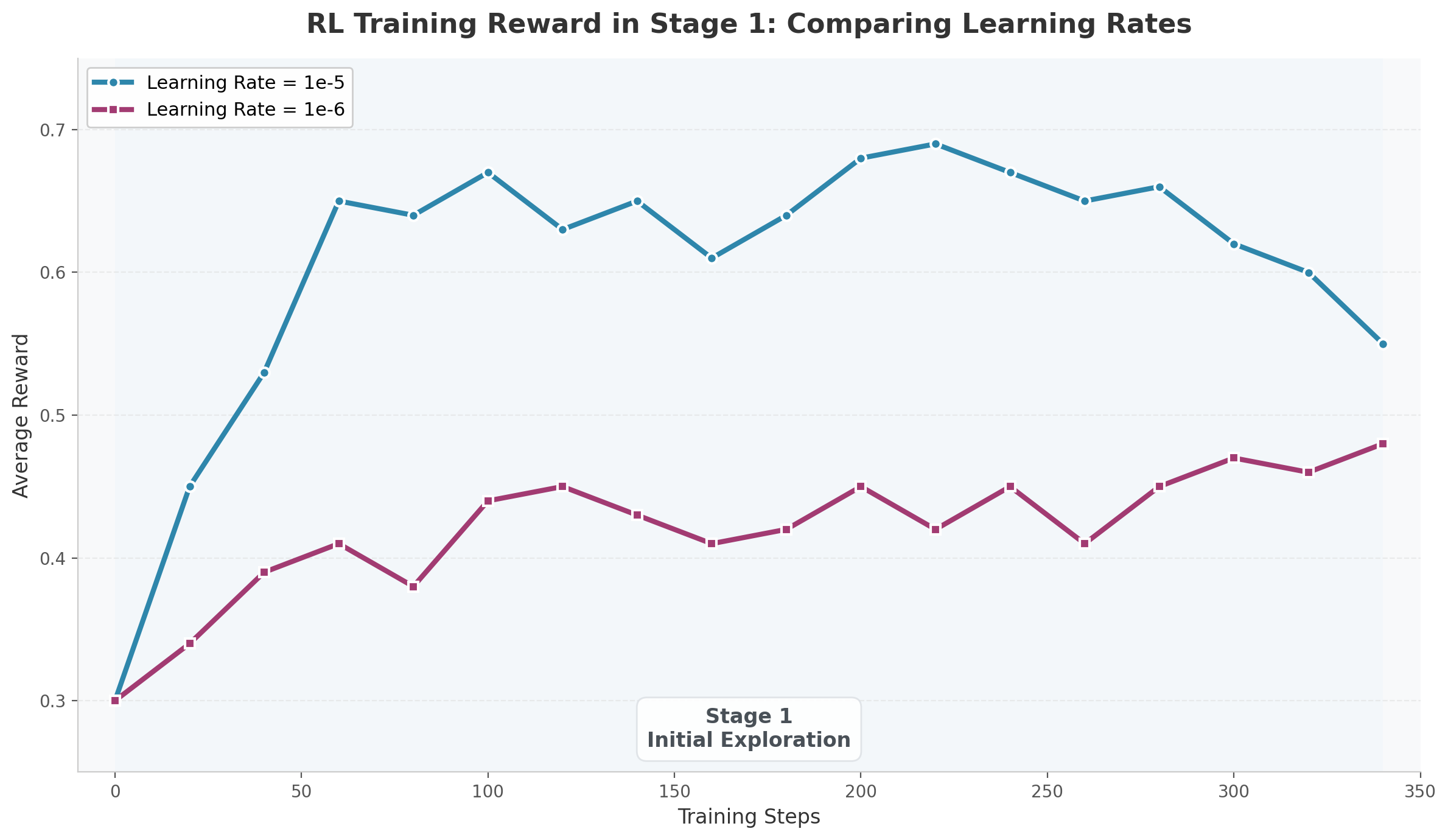}
        \caption{Effect of Learning Rate on Training Reward}
        \label{fig:curriculum_learning_lr}
    \end{subfigure}
  \caption{
    \textbf{Comparison of Curriculum Learning Aspects.}
    This figure analyzes key aspects of curriculum learning.
    \textbf{(a)} illustrates the significant drop in validation accuracy during Stage 2.
    \textbf{(b)} shows that a higher learning rate can lead to faster learning efficiency but also instability during Stage 1.
}
    \label{fig:combined_curriculum_learning}
\end{figure}

We attempted to introduce a staged curriculum learning strategy, aiming to construct a reinforcement training process that progresses from easy to difficult. Specifically, we began by using a medium-difficulty math problem set (such as regular problems covering K12 knowledge, defined as "Normal Problems") for initial training to obtain a model checkpoint at this stage. Then, using this checkpoint as the starting point, we switched to a high-difficulty problem set (such as competition questions, defined as "Hard Problems") to continue training, with the goal of achieving progressive enhancement in reasoning ability.

However, the experimental results (Figure \ref{fig:combined_curriculum_learning}) indicate that difficulty-based data switching failed to improve generalization, likely due to distribution shift. Specifically, after switching to the Hard Problems for training, although the model's adaptation to complex tasks improved (with stage 2 reward values continuously increasing), its performance on the unified evaluation set showed divergence: the accuracy on Normal problems significantly decreased, and the performance on some logic and physics subtasks fluctuated or stagnated. This phenomenon indicates that the complex skills, special patterns, or high-level strategies learned by the model on the Hard Problems, while helpful for tackling high-difficulty tasks, might conflict with the core reasoning paths relied upon by medium-difficulty problems, ultimately weakening its overall generalization ability.

\subsection{Learning Rate Strategy of RL} 
\label{sec: learning rate strategy}
As depicted in Figure \ref{fig:curriculum_learning_lr}, our initial experiments demonstrate that the learning rate significantly impacts the efficiency of RL, as the higher. This hyperparameter dictates the magnitude of each weight update during optimization. We define a higher learning rate as 1e-5 and a lower rate as 1e-6. Although higher rates typically enable a faster initial reward gain in traditional ML (consistent with our Stage 1 results), they introduce substantial risks of instability later in training. Specifically, Stage 1 revealed that 1e-5 accelerates early reward accumulation but can precipitate model collapse in subsequent phases, manifesting itself as reward diminishment and entropy collapse.

\begin{figure}[htbp]
    \centering
    \begin{subfigure}
    {0.48\textwidth}
        \centering
        \includegraphics[width=\linewidth]{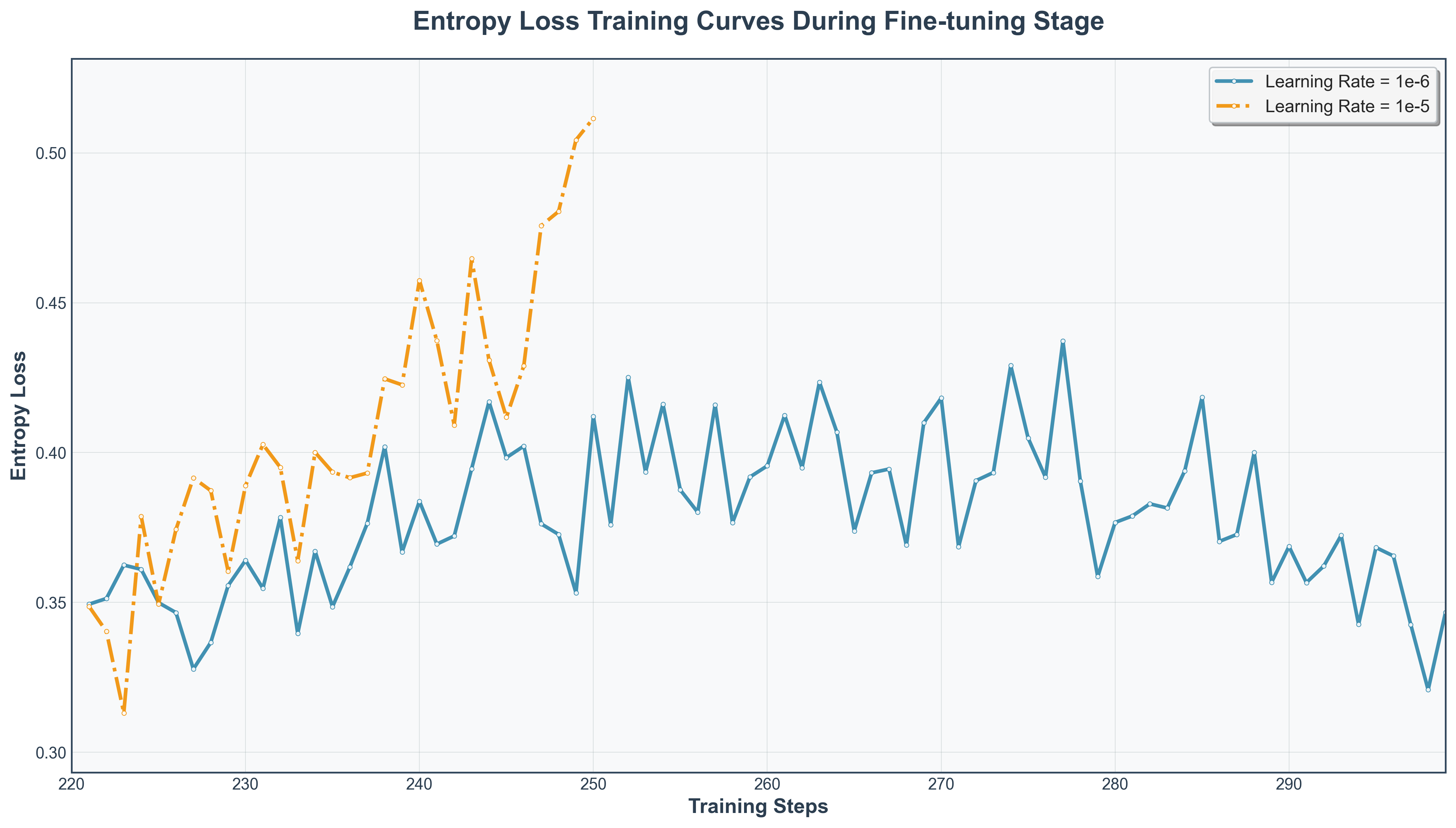}
        \caption{Entropy Loss during Stage 2}
        \label{fig:lr_entropy}
    \end{subfigure}
    \hfill
    \begin{subfigure}{0.48\textwidth}
        \centering
        \includegraphics[width=\linewidth]{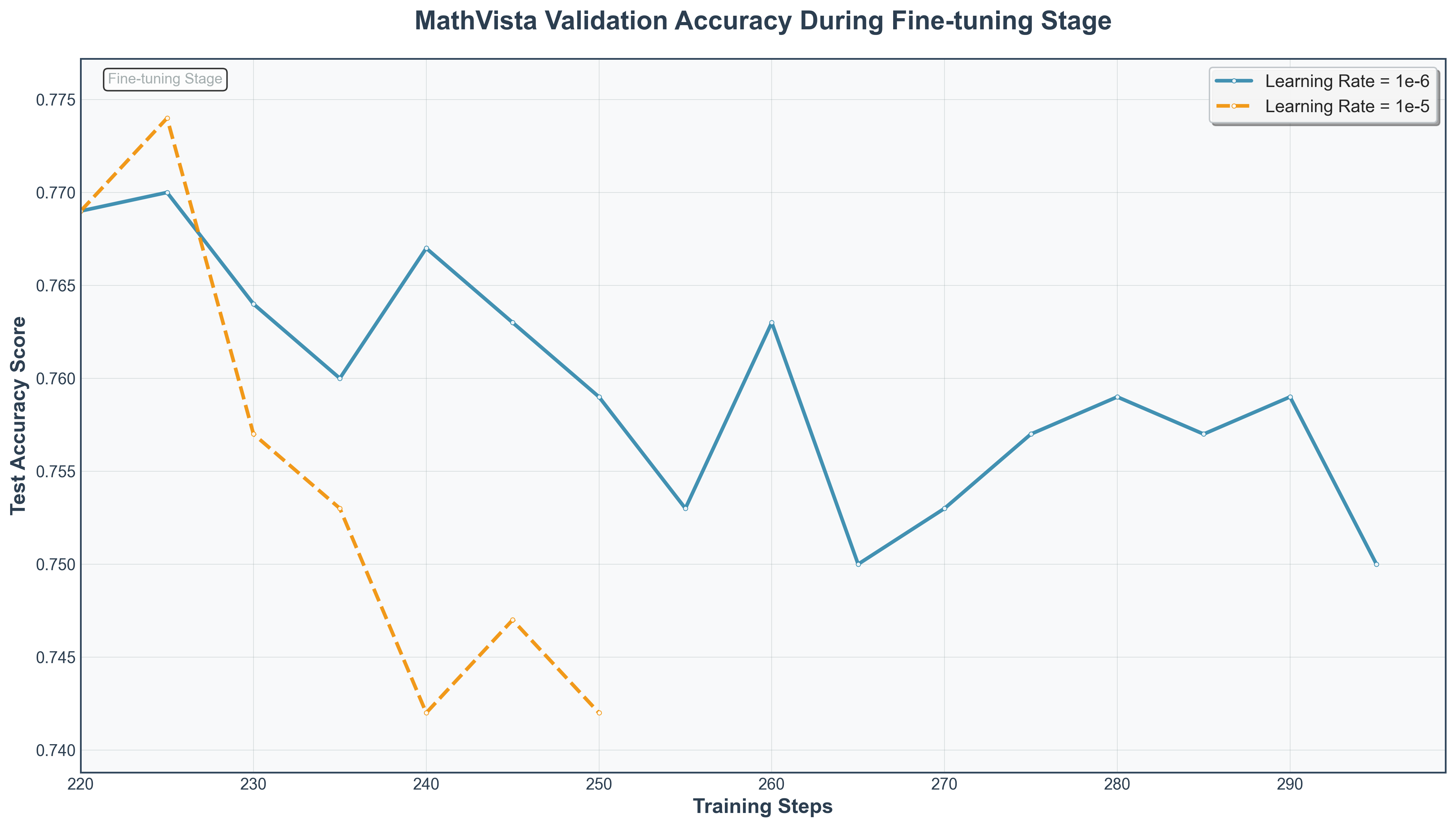}
        \caption{MathVista Score during Stage 2}
        \label{fig:lr_acc}
    \end{subfigure}
    \caption{
        \textbf{Impact of Learning Rates on RL Performance in Stage 2 (Hard Problems).}
        This figure shows key metrics for different learning rate strategies on Hard Problems
        \textbf{(a)} A higher learning rate indicates the policy's instability on exploration when tackling Hard Problems.
        \textbf{(b)} This indicates the model struggle to generalize from harder to easier problems.
    }
\end{figure}

Conversely, Figure \ref{fig:lr_entropy} and \ref{fig:lr_acc} show that in Stage 2, maintaining a learning rate of 1e-5 after initial convergence led to a critical observation: the model's entropy quickly diverged from its stabilized trend and exhibited a sharp increase in the log metrics. This unexpected surge signifies policy destabilization – the model effectively "unlearns" its acquired behavior due to the overly aggressive updates. The large learning rate prevents sustained finetuning and jeopardizes the stability of the converged policy. Consequently, while 1e-5 enhances initial learning speed, the significant risk of late-stage instability necessitates careful consideration for efficiency and final performance.

\subsection{Additional Reinforcement Learning Trials}
\label{sec: extra_rl}
In our experimental setup, we also explored two approaches: Clip-Higher and Dynamic Sampling.

\paragraph{Dynamic Sampling} In our experiments, dynamic sampling moderately impacts our baseline model. Specifically, it mitigates diminishing training progress, stabilizes the learning curve for extended training steps, yet ultimately fails to improve validation set performance.

\paragraph{Clip-Higher} 

Clip-Higher was primarily intended to mitigate pattern convergence, yet it's not performing as expected. We're observing severe instability and anomalous gradient norms during training. We hypothesize this aligns with the covariance-driven entropy dynamics described by \cite{cui2025entropymechanismreinforcementlearning}: entropy change is governed by the negative covariance between action log-probabilities and advantage-driven logit updates. Specifically, when high-advantage tokens coincide with low initial probabilities (acting as exploration candidates), their gradient updates exhibit a large positive covariance. Clip-Higher, by permitting larger probability increments, amplifies this effect. While this amplification can boost exploration, when combined with high learning rates, it risks over-amplifying noisy advantages and destabilizing the loss landscape. This, in turn, leads to severe loss explosions and ultimately, model collapse.

% \paragraph{KL Loss of GRPO} Traditional RLHF like PPO utilize the KL divergence to mitigate the overoptimization, which is directly applied to 

\subsection{Effectiveness of Connector-Only Tuning After RL Stage}\label{mlp_after_rl}

After undergoing RL training on mathematical data, the model demonstrates improved reasoning ability. To further enhance its generalization, we apply Connector-Only Tuning with multi-domain knowledge data to re-balance the knowledge base and achieve better performance across a wide range of domains.  Here, we conduct an ablation study to investigate the effect of tuning different components (ViT/LLM/Connector) in this stage.

As summarized in Table~\ref{tab:connector_tuning_horizontal}, tuning the LLM or the ViT components alongside the connector significantly degrades performance, with accuracy dropping to 73.2 and 69.7 respectively. This suggests that excessive parameter updates in these core components disrupt the reasoning patterns reinforced during RL. In contrast, Connector-Only Tuning, which leaves the model’s intrinsic language and vision representations untouched, effectively preserves the reasoning capability while incorporating multi-domain knowledge. Consequently, Connector-Only Tuning significantly improves the model’s performance on the MMMU benchmark, achieving an accuracy of 76.0. This phenomenon further demonstrates the fundamental role of the connector module in facilitating robust cross-modal alignment in multimodal reasoning models.

\begin{table}[ht]
\centering
\caption{Ablation study on tuning different components after RL stage (MMMU benchmark).}
\label{tab:connector_tuning_horizontal}
\begin{tabular}{lcccc}
\toprule
 & \makecell{\textbf{RL Stage} \\ (baseline)} 
 & \makecell{\textbf{Connector-Only} \\ \textbf{Tuning}} 
 & \makecell{\textbf{Connector +} \\ \textbf{LLM Tuning}} 
 & \makecell{\textbf{Connector + ViT} \\ \textbf{+ LLM Tuning}} \\
\midrule
Accuracy (\%) & 74.5 & \textbf{76.0} & 73.2 & 69.7 \\
\bottomrule
\end{tabular}
\end{table}

% \section{Curriculum Learning in RL}
% exploration in RL

% \section{Observation on }
% Data can 

\vspace{-2mm}
\section{Discussions}
In the following, we conducted a series of carefully designed exploratory experiments to better understand the nature of reasoning in visual language models.

\subsection{Exploring Whether Visual Language Models Develop Reasoning via Memorization or Generalization}
\label{sec: memory_or_general}

In practice, we observe that some models are able to produce responses with a reasoning-like style and achieve relatively high scores on in-distribution validation sets. However, in most cases, such behavior merely reflects memorization of existing patterns rather than genuine activation of a reasoning ability that generalizes across domains. These models tend to generate outputs with a \textit{reasoning style} without truly acquiring a \textit{generalizable reasoning ability}. The following experiments illustrate this distinction.

We take the cold-start model as the baseline and compare two different training strategies with  \textbf{mathematics data}: RL (implementation details refer to Section~\ref{sec:rl}) and SFT (conducted with a global batch size of 64 and a learning rate of $1\mathrm{e}{-5}$, scheduled using a cosine decay with a warmup ratio of 0.03) using reasoning-style data distilled from our Skywork-R1V3-38B. We evaluate both models on \textit{in-domain} (MathVista) and \textit{out-of-domain} (MMMU) benchmarks to examine their ability to generalize beyond the domain of training.

\begin{table}[ht]
\centering
\caption{Comparison of memorization and generalization between RL and SFT.}
\label{tab:memorization_vs_generalization}
\begin{tabular}{lccc}
\toprule
\textbf{Method} & \textbf{In-Domain } & \textbf{Out-of-Domain} & \textbf{Reasoning Style} \\
\midrule
 Baseline & 71.3 & 68.1 & - \\
RL                  & \textbf{77.2} & \textbf{74.5} & Yes \\
SFT                 & 75.9 & 65.4 & Yes \\
\bottomrule
\end{tabular}
\end{table}

As shown in Table~\ref{tab:memorization_vs_generalization}, both RL and SFT enhance the model’s in-domain mathematical reasoning ability and result in responses with reasoning-like style. However, only RL leads to improved out-of-domain performance (74.5 on MMMU), demonstrating that the reasoning ability acquired through RL is genuinely generalizable. In contrast, SFT fails to generalize and even harms the out-of-domain performance (dropping to 65.4), likely due to overfitting to biased knowledge introduced at the finetuning stage. These results highlight that RL effectively activates a true generalizable reasoning ability, while SFT tends to merely imprint a reasoning style on the outputs without substantial generalization.

\subsection{Exploration on Slow-thinking to Fast thinking in Inference}
\label{sec: fast_to_slow}
% \subsection{Chat template Design}

\paragraph{Slow-Thinking} We adapted the following chat template in slow-thinking mode. The structure follows a turn-based format:

\begin{lstlisting}
<|im_start|>system
You are a helpful assistant.<|im_end|>
<|im_start|>human
[user input]<|im_end|>
<|im_start|>assistant
<think>[reasoning process]</think>
[summary response]<|im_end|>
\end{lstlisting}
% \hfill
% \hfill
% \hfill

\paragraph{NoThink Mode} 
We adapted the following chat template to explicitly activate the fast-thinking mode. The key distinction between this template and the slow thinking template lies in the inclusion of the \texttt{<think>} token after the \texttt{assistant} role token. Removing this token prompts the model to output an answer directly without engaging in intermediate reasoning. We denote the inference without \texttt{<think>} token as NoThink mode. 
\begin{lstlisting}
<|im_start|>system
You are a helpful assistant.<|im_end|>
<|im_start|>human
[user input]<|im_end|>
<|im_start|>assistant
[response]<|im_end|>
\end{lstlisting}

\paragraph{NoWait Mode} 
Following~\cite{wang2025waitdontneedwait}, we implement another fast-thinking method by constraining the generation during inference. Specifically, we prohibit certain deliberative tokens such as \texttt{Wait}, \texttt{Alternatively}, and \texttt{Hmm} via logit manipulation. This strategy suppresses the tendency toward exploratory reasoning and encourages immediate answer production, denoted as \textbf{NoWait} mode.

%The key distinction between the NoThink mode and our original mode lies in the inclusion of the \texttt{<think>} token after the \texttt{assistant} role token. By toggling this token, the model system seamlessly switches between slow mode (with explicit reasoning steps) and fast mode (direct response generation).
%We denote inference using the above fast-thinking template on the original Skywork-R1V-38B as  NoThink.

\begin{table}[ht]
\centering
\caption{Comparison of slow-thinking and fast-thinking inference modes on the MMMU benchmark.}
\label{tab:slow_fast_thinking_horizontal}
\begin{tabular}{lcccc}
\toprule
 & \textbf{Slow Thinking} & \textbf{NoThink} & \textbf{NoWait} \\
\midrule
Avg Tokens & 1272  & 263 & 1040 \\
Accuracy (\%) & 76.0  & 68.9 & 74.1 \\
\bottomrule
\end{tabular}
\end{table}

Table~\ref{tab:slow_fast_thinking_horizontal} presents a comparison of slow-thinking and fast-thinking inference modes on the MMMU benchmark. The results reveal a clear trade-off between response length and reasoning performance. The \textit{Slow Thinking} mode achieves the highest accuracy (76.0\%), benefiting from its explicit step-by-step reasoning process, albeit at the cost of generating significantly longer outputs (1272 tokens on average). In contrast, \textit{NoThink}, which removes the reasoning phase entirely, produces extremely concise responses (263 tokens) but suffers a substantial drop in accuracy (68.9\%). \textit{NoWait}, which enforces a more direct answering style by banning deliberative tokens during inference, strikes a better balance: it reduces the average output length by approximately 18\% compared to \textit{Slow Thinking} while retaining a relatively high level of accuracy (74.1\%).

\subsection{The Effectiveness of Thinking Budget}
\label{sec: thining budget}
\begin{figure}[!ht]
    \centering
    \includegraphics[width=0.65\linewidth]{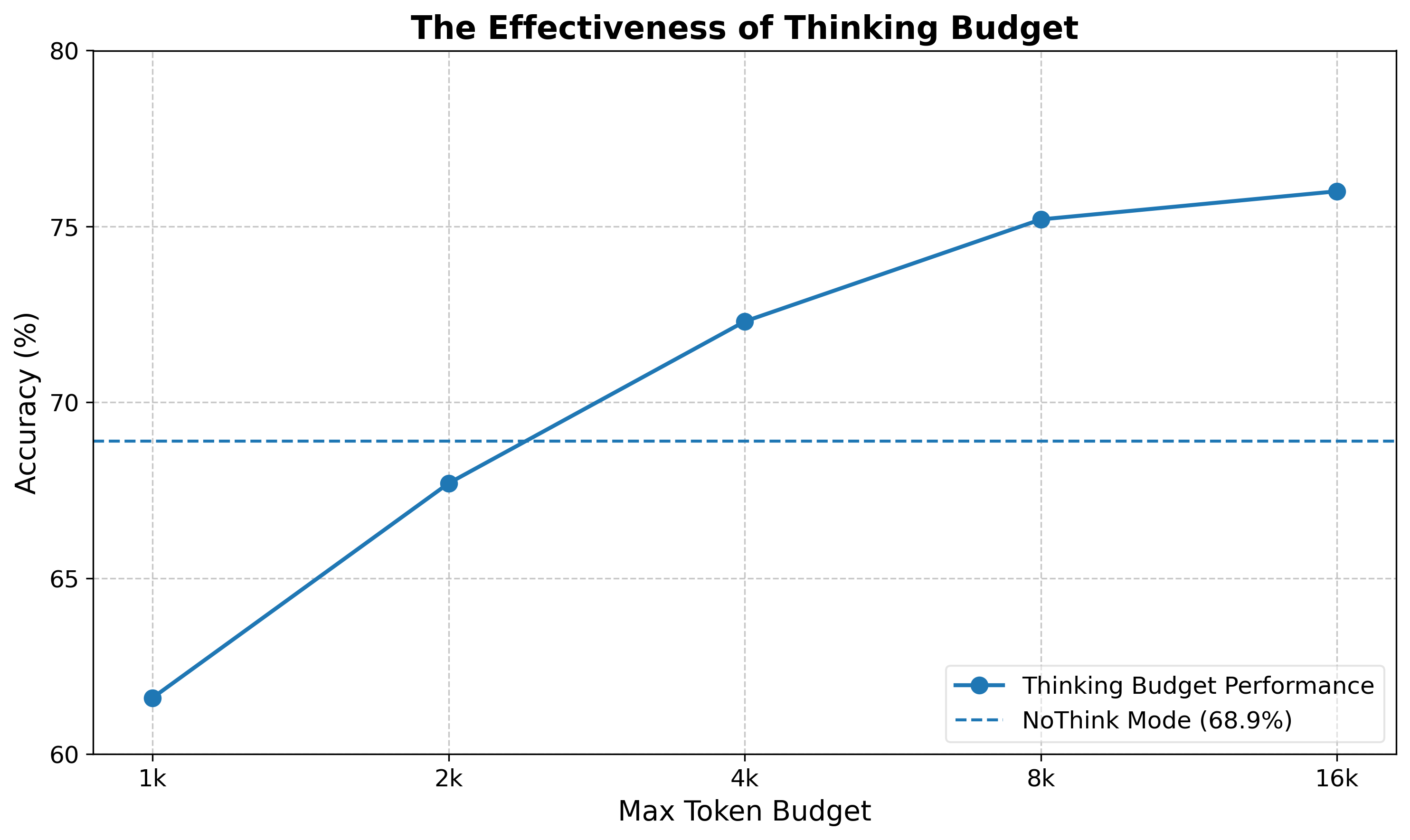}
    \caption{Model performance with different thinking budgets.}
    \label{fig:think_budget}
\end{figure}

We further investigate how the allocated \textit{thinking budget}, defined as the maximum token allowance during generation, affects the model’s reasoning performance. The results, summarized in Figure~\ref{fig:think_budget}, reveal a test-time scaling law: reasoning accuracy on the MMMU benchmark improves steadily with increased token budgets. 

As shown in the figure, the accuracy on the MMMU benchmark increases steadily with a larger thinking budget. Specifically, allocating only 1,024 tokens yields an accuracy of 61.6\%, which already lags behind the \textit{no-thinking} mode baseline (68.9\%). Increasing the budget to 2K tokens closes this gap, reaching 67.7\%. Further increasing the budget to 4K and 8K tokens leads to significant gains, with accuracies of 72.3\% and 75.2\%, respectively. When the budget is extended to 16K tokens, the performance saturates at 76.0\%, achieving the highest observed accuracy. These results demonstrate a clear trade-off between computational cost and reasoning effectiveness. Allowing a sufficient token budget is critical to fully leverage the model’s step-by-step reasoning capabilities.

\subsection{Hallucination in Skywork-R1V3's Chain-of-Thought Impairs Reasoning Performance}
\label{sec: hallucination}
\begin{table}[ht]
\centering
\caption{Impact of ``I can't see the image'' on MMMU Performance}
\label{tab:mmmu_image_issue}
\begin{tabular}{lcc}
\toprule
\textbf{Subset} & \textbf{Number of Samples} & \textbf{Accuracy (\%)} \\
\midrule
All & 900 & 76.0 \\
Without ``I can't see the image'' & 853 & 77.3 \\
With ``I can't see the image'' & 47 & 51.1 \\
\bottomrule
\end{tabular}
\end{table}

We observe an intriguing phenomenon in Skywork-R1V3: hallucinated statements like "can't see the image" frequently appear within chain-of-thought (CoT) rationales but are consistently absent from the final summary responses. This inherited behaviour—first noted in Skywork-R1V2, which reveals that despite explicitly stating an inability to perceive images within its reasoning trace, the model's subsequent rationale demonstrates clear understanding of the visual content. This paradox represents one of the first documented challenges in transferring text-based reasoning capabilities to vision-language settings.

Critically, we found that the presence of this "can't see the image" hallucination within the CoT significantly impedes model performance in downstream evaluations, as shown in Table \ref{tab:mmmu_image_issue}. Our detailed analysis suggests this effect stems from the hallucination disrupting the coherence of the reasoning process. While the phenomenon may also relate to the entropy of crucial reasoning tokens, we leave further investigation of this mechanism as an open question for the community.

% \subsection{Curriculum Learning in RL}
% \wshen{write something about the curriculum learning in RL}

\subsection{Analysis on Entropy Token in Visual Reasoning Task}
\label{entropy tokens analysis}

\begin{figure}[!h]
    \centering
    \includegraphics[width=0.6\linewidth]{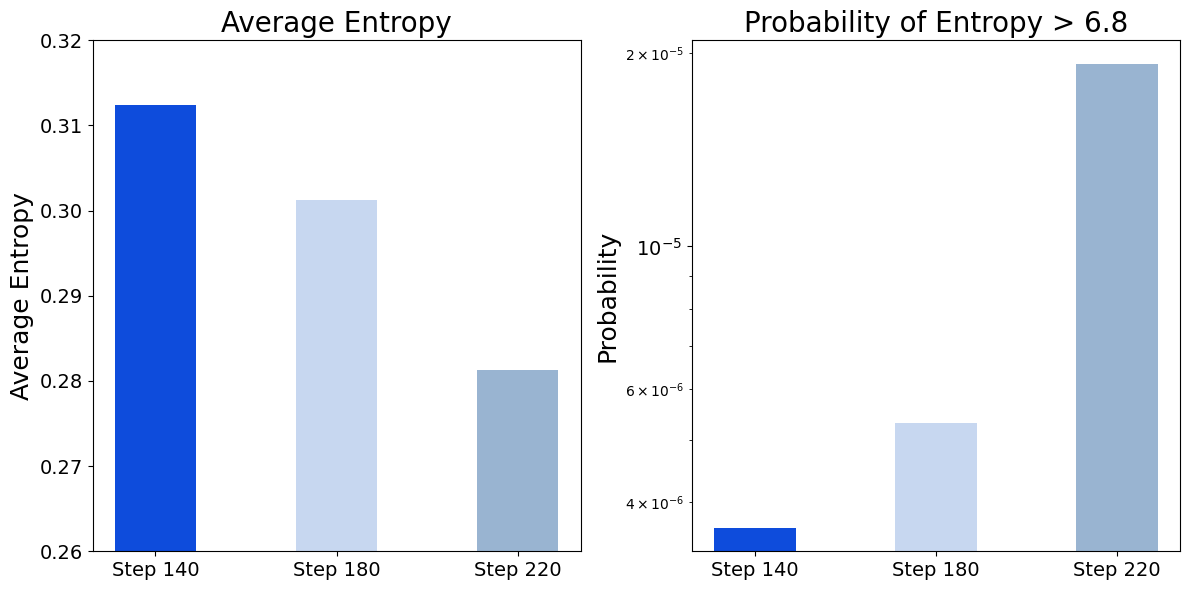}
    \caption{Entropy distribution at different GRPO steps}
    \label{fig:Entropy_Distribution}
\end{figure}

We observed an intriguing phenomenon during the RL stage when the reward steadily increased: while the average entropy of the generated tokens (i.e., the mean entropy of the vocabulary probability distribution  at each forward step) progressively decreased, the probability of encountering high-entropy samples, although small, actually increased.

Figure~\ref{fig:Entropy_Distribution} (left) illustrates the average entropy at different reinforcement training stages. It can be clearly seen that as training steps progress from 140 to 220, the average entropy consistently declines. However, Figure\ref{fig:Entropy_Distribution}~(right) indicates that the probability of generating tokens with high entropy (defined here as entropy greater than 6.8) notably increases at later training stages. This counterintuitive observation suggests that although the model generally becomes more deterministic in its predictions through reinforcement learning, it also occasionally exhibits greater uncertainty, possibly reflecting increased exploration or exposure to challenging samples that require more deliberate reasoning \citep{wang20258020rulehighentropyminority}.

\subsection{The Limitations of Self-Incentivized Reasoning for Instructed Models in RL Training}
\label{sec: self-incentivized reasoning}
Test time scaling law \citep{jaech2024openai} strongly suggests that extended reasoning traces ("thinking before answering") enhance model performance on complex tasks. However, the field lacks robust visual reasoning models capable of sustaining such processes. Two dominant approaches exist for encouraging longer reasoning traces during training: (1) self-incentivized methods where models generate verification steps through instructional prompts (e.g., ``\texttt{The reasoning process MUST BE enclosed within tags. The final answer MUST BE put in \textbackslash boxed{}.}''), and (2) cold-start data priming prior to RL optimization.

\begin{figure}[!h]
    \centering
    \includegraphics[width=0.8\linewidth]{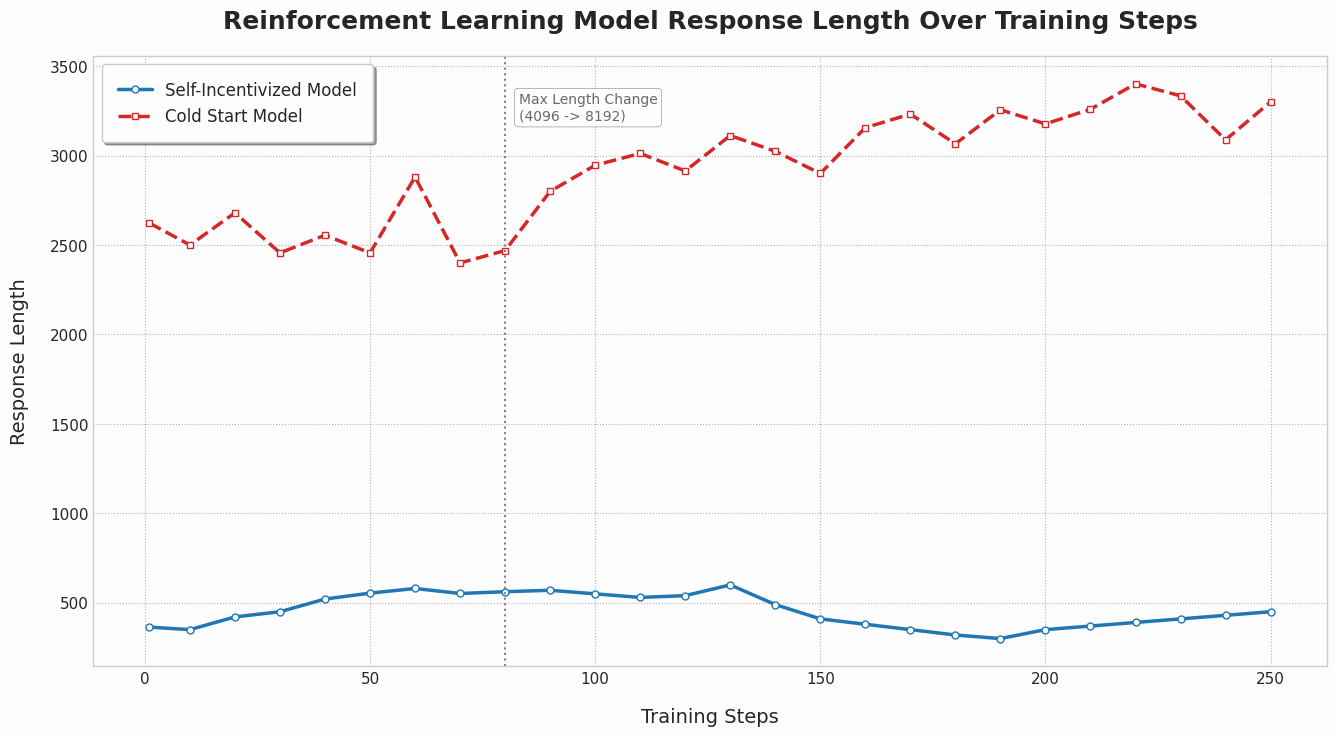}
    \caption{Response length during training: A comparison of two base models}
    \label{fig:length_change}
\end{figure}
Our experiments reveal a critical limitation of the self-incentive approach. As shown in Figure \ref{fig:length_change}, models fail to develop genuinely long Chains of Thought (CoT) from scratch, exhibiting neither sustained "aha moments" nor length expansion proportional to reward growth before collapsing in later RL stages. This stands in stark contrast to cold-start initialization, which enables progressive reasoning trace elongation and significantly boosts both training and evaluation rewards. These findings align with recent work like DeepSeek-R1 and suggest that instruction-tuned models may suppress critical reasoning tokens (e.g., "Wait," "Alternatively") during finetuning—precisely those tokens that diversify reasoning paths and increase solution probability. The emergence of "aha moments" appears fundamentally linked to these suppressed tokens. Notably, RL-Zero approaches \citep{hu2025openreasonerzeroopensourceapproach} demonstrate success comparable to cold-start methods, indicating that the specific RL initialization point is less critical than ensuring models can access and amplify critical tokens during their thinking process.

% This presents a core challenge for developing strong reasoning models: we must efficiently teach models to (1) recognize and (2) utilize critical reasoning tokens during their pre-answer deliberation. 
% Until open-source visual reasoning models mature, a promising interim solution involves using stitched vision models to distill valid vision-specific long CoT examples.

% \subsection{Does Qwen based model also work when exposed to random reward within the visual-related task?}

% \subsection{Dose mitigate overthinking can really improve the reasoning efficiency?}

\section{Conclusion}
\label{sec:conclusion}
Skywork-R1V3 represents a groundbreaking advance in open-source vision-reasoning models (VRMs), establishing a new multimodal reasoning benchmark through its reinforcement learning (RL)-based training paradigm. As the first predominantly RL-enhanced open-source VLM, it achieves exceptional visual reasoning capabilities without external pretraining data while demonstrating remarkable generalization through cross-domain transfer—particularly from mathematical to other subject domains.

Our work showcases RL's transformative potential for refining vision-language alignment and provides crucial insights into cross-modal reasoning transfer. By seamlessly integrating textual and visual understanding, Skywork-R1V3 establishes the foundation for scalable, generalizable multimodal AI systems. This advancement marks a major milestone in open-source VLM development, offering a robust framework for RL-driven multimodal intelligence.

\section{Limitations and Future Work}
\paragraph{Limitations}
The scope of this report is intentionally narrow, and as such, we do not explore the specifics of vision encoder architectures or large language model (LLM) pre-training. Consequently, enhancing general visual capabilities was not a primary objective of this work. Looking ahead, we identify meticulous data curation as a critical direction for future research. We believe a more rigorously curated dataset will be instrumental in developing robust and capable models. Additionally, further improvements are needed in Video Understanding and Agent Training.

Another limitation we must acknowledge is the hallucination issue inherent in the reasoning contents of R1V2. We observed that this can hinder efficient reasoning and occasionally lead to erratic outputs. While this version of the reasoning model does not negate the benefits of reinforcement learning (RL), we plan to address this issue in future work.

% \wshen{1. cant see the image problem. hallucination from the R1V
% 2. answer the question}
\paragraph{Integrating Tool Use into Agentic Vision-Language Models with Reinforcement Learning}
The recent emergence of models such as OpenAI o3 and o4-mini highlights the growing potential of VLMs to perform tool-augmented reasoning. Such models extend beyond single-turn inference of Skywork-R1V3 by incorporating operations like code execution, web browsing, and image manipulation (e.g., cropping and zooming) to process and iteratively interact with visual inputs. Building upon this progress, a pivotal direction for future research lies in equipping VLMs with the ability to invoke external tools to address complex, multi-step multimodal tasks. In particular, end-to-end reinforcement learning approaches—such as Reinforcement Learning with Verifiable rewards (RLVR) \citep{mroueh2025reinforcementlearningverifiablerewards}—offer a promising paradigm for allowing models to acquire tool-use skills through environmental interaction. 
However, their development faces significant challenges due to system optimization complexity and prohibitive inference computational costs, presenting a critical research gap that warrants further investigation.

\paragraph{Towards to Unified Vision-Language Models}
While this work focuses on enhancing reasoning capabilities through targeted training strategies, an important future direction is the unification of visual understanding and generation within a single VLM. A unified architecture, jointly trained on understanding and generation tasks, has the potential to leverage complementary signals, enhance multimodal alignment, and improve generalization across diverse scenarios~\citep{zhang2025unifiedvisionlanguagemodelsnecessary}. Pursuing this direction may provide valuable insights into the design of VLMs capable of seamlessly transitioning between comprehension and generation, thereby expanding their utility in real-world applications that demand both modalities.

\paragraph{Advancing Physics Reasoning and Embodied Agents for Real-World Comprehension}
While Skywork-R1V3 demonstrates strong multimodal reasoning capabilities, particularly in STEM domains like mathematics and physics, further progress requires deepening its understanding of physical dynamics within embodied contexts.
Enhancing the model's physics reasoning will establish foundations for advanced embodied intelligence. We specifically aim to integrate such models into agentic systems capable of perceiving and interacting with simulated or physical environments. This necessitates not only high-fidelity vision-language alignment but also temporal planning and adaptive learning through interaction. Online reinforcement learning provides a promising pathway for agents to acquire complex reasoning and tool-use abilities via environmental feedback loops. Such experiential learning yields physical grounding that significantly improves real-world applicability. Furthermore, future agents must process multimodal sensory inputs in real-time and operate within physical constraints to ensure generalization and robustness.
These advances will ultimately enable autonomous systems that understand and reason about the physical world with human-like competence and reliability.

\section{Contributions}
\paragraph{Core Contributors:}
Wei Shen*, Jiangbo Pei, Yi Peng, Xuchen Song\textsuperscript{\textdagger}, Yang Liu

\paragraph{Contributors:}
Jian Peng, Haofeng Sun, Yunzhuo Hao, Peiyu Wang, Jianhao Zhang, Yahui Zhou

\vspace{0.2cm}
* \href{mailto:wei.shen@kunlun-inc.com}{wei.shen@kunlun-inc.com}
\textsuperscript{\textdagger}Project Leader \quad

\clearpage
\appendix

\clearpage
\bibliography{main
}
\bibliographystyle{colm2024_conference}

\newpage

\appendix

\section*{\hspace{-4mm} \centering Appendix}
\vspace{3mm}

% \section{Contributions}
% \paragraph{Core Contributors:}
% Wei Shen*, Jiangbo Pei, Yi Peng, Xuchen % Song\textsuperscript{\textdagger}, Yang Liu

% \paragraph{Contributors:}
% Jian Peng, Haofeng Sun, Yunzhuo Hao, Peiyu Wang, Yahui Zhou

% \vspace{0.2cm}
% * \href{mailto:wei.shen@kunlun-inc.com}{wei.shen@kunlun-inc.com}
% \textsuperscript{\textdagger}Corresponding author \quad

% \paragraph{Core Contributor:} 
% \quad Wei Shen \quad Jiangbo Pei \quad Yi Peng \\ 
% \quad Xuchen Song* \quad Yang Liu*

% \paragraph{Contributor:} 
% \quad Jian Peng \quad Haofeng Sun \quad Yunzhuo Hao \\ 
% \quad Peiyu Wang \quad Yahui Zhou

% \input{acknowledge}

% \section{Acknowledgment}
%We're grateful to JianHao Zhang for their AI infrastructure support, and to Xiaokun Wang, Zifeng Cao and Cheng Yu for their insightful suggestions. We also acknowledge the contributions of the Skywork-R1V2 authors.

%Our deepest thanks go to Mr. Yahui Zhou, founder of Kunlun Inc., whose indispensable financial support for the R1V-series work and providing access to GPU resources was crucial for this study's successful completion.

%\newpage

\section{Qualitative Examples}
\begin{figure}[H]
    \centering
    \includegraphics[width=0.25\textwidth]{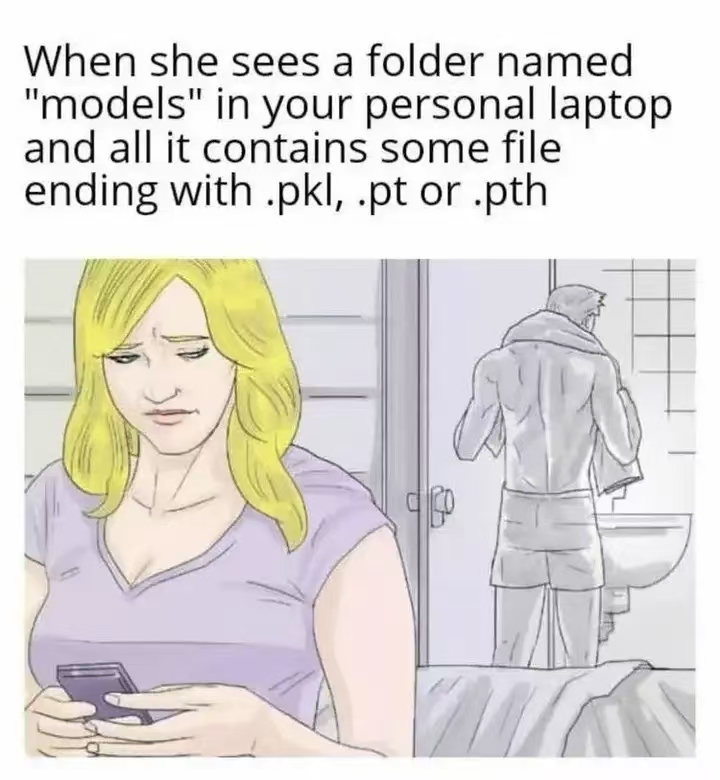}
    \caption{example 1}
    \label{fig:example}
\end{figure}
\begin{CJK}{UTF8}{gbsn}
\begin{tcolorbox}[title=Prompt]
请解释这个 MEME 的含义
\end{tcolorbox}
\begin{tcolorbox}[title=Model Response]
\verb|<think>|
好的，我现在需要分析用户提供的这个 meme。首先看一下图片和文字内容。图片上有两个人物：前面是一位女士坐在床边，低头看着手机，表情有些困惑或不悦。后面是一个男人站在洗手间门口，似乎正在换衣服或者整理衣物。文字部分写的是：“当她在你的个人笔记本电脑里看到一个名为“models”的文件夹，里面包含一些以 .pkl、.pt 或 .pth 结尾的文件。”

首先，我需要理解各个元素的含义。文件夹名 “models” 在编程中通常用来存放机器学习模型的文件。而 .pkl 是 Python Pickle 文件的扩展名，用于保存任意 Python 对象；.pt 和 .pth 则是 PyTorch 框架中常用的模型保存格式。所以这里的 “models” 文件夹实际上是指机器学习相关的模型文件。

接下来，为什么这会引发女士的不快呢？可能是因为 “models” 这个词在日常生活中也可能指代模特或与之相关的内容。女士可能误解了文件夹的用途，以为里面存放的是与模特相关的东西（比如照片或个人信息），从而产生怀疑或调侃的情绪。而实际上，这些文件是技术性的，和机器学习模型有关。

另外，图片中男士的动作可能暗示他正在逃避或回避女士的质问，进一步加强了这种误解带来的幽默感。这种反差（技术术语与日常生活中的误解）是这个 meme 的核心笑点。

需要确认是否存在其他可能的解释，但结合常见的编程术语和日常生活中的双关语，上述分析应该是正确的。最后，确保回答清晰易懂，解释技术和幽默之间的联系。
\verb|</think>|

这个 meme 的幽默点在于技术术语与日常生活中的误解之间的反差，具体解析如下：

\begin{itemize}
    \item \textbf{技术背景:} \\
    文件夹名 “models” 及扩展名 .pkl、.pt、.pth，都是机器学习中的常见术语：\\
    - .pkl 是 Python 用于保存对象的格式。\\
    - .pt 和 .pth 是 PyTorch 框架保存模型的格式。\\
    这里的 “models” 指代的是机器学习模型文件，而非字面意义的“模特”或“模型”（如人工制品）。
    \item \textbf{日常生活中的误解:} \\
    图片中的女性看到这些文件后，可能误以为 “models” 文件夹与现实中展示身材的 “模特” 相关（例如私密照片或不忠的证据），从而产生怀疑或调侃的情绪。而实际上这些文件只是技术文档，这种误会形成了笑点。
    \item \textbf{画面的补充:} \\
    男性站在浴室门口、似乎在整理衣物的动作，强化了女性对“另一段关系”的联想，让情境更显滑稽。
\end{itemize}

\textbf{总结:} 这个 meme 利用了 “models” 一词在编程与日常用语中的双重含义，通过技术细节与生活场景的对比，制造了幽默效果。
\end{tcolorbox}

\noindent
\begin{minipage}{\textwidth}
    \begin{minipage}{0.3\textwidth}
        \centering
        \includegraphics[width=0.95\linewidth]{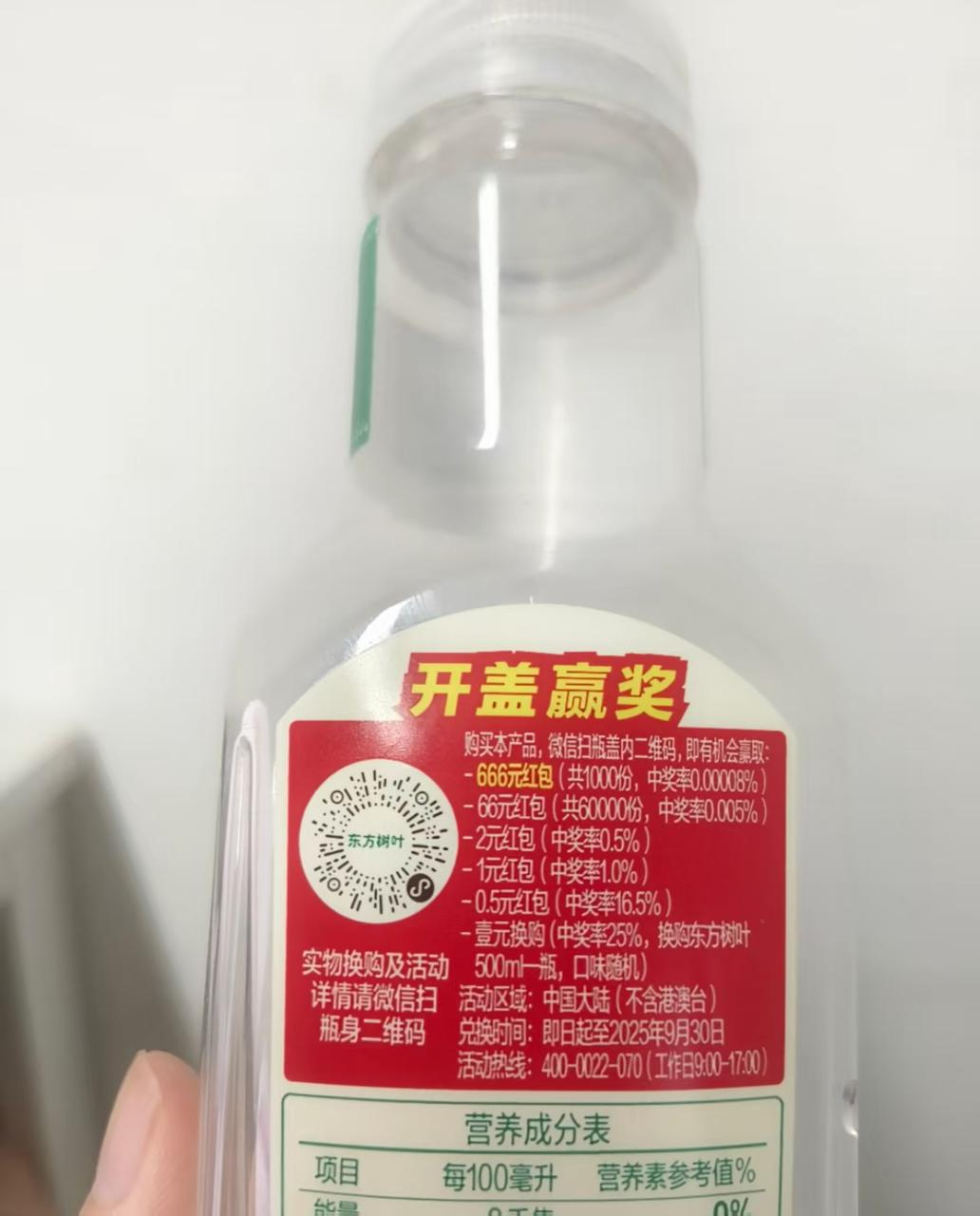}
        \captionof{figure}{example 2}
        \label{fig:tea}
    \end{minipage}
    \hfill
    \begin{minipage}{0.65\textwidth}
        \begin{tcolorbox}[title=Prompt]
        我买一千瓶这个饮料，中奖的钱的期望是多少？
        \end{tcolorbox}

        \begin{tcolorbox}[title=Model Response]
        \verb|<think>|...\verb|</think>|\\
        根据瓶身标签信息，需要计算各奖项的中奖金额期望值。具体如下：

        \vspace{0.5em}
        \textbf{各奖项及其中奖概率：}
        \begin{itemize}
          \item 666 元红包：0.00008\%（即 $0.0000008$）
          \item 66 元红包：0.005\%（即 $0.00005$）
          \item 2 元红包：0.5\%（即 $0.005$）
          \item 1 元红包：1.0\%（即 $0.01$）
          \item 0.5 元红包：16.5\%（即 $0.165$）
        \end{itemize}

        \textbf{各奖项期望值：}
        \begin{itemize}
          \item 666 元：0.0005328 元/瓶
          \item 66 元：0.0033 元/瓶
          \item 2 元：0.01 元/瓶
          \item 1 元：0.01 元/瓶
          \item 0.5 元：0.0825 元/瓶
        \end{itemize}

        汇总：
        \[
        0.0005328 + 0.0033 + 0.01 + 0.01 + 0.0825 = 0.1063\,\text{元/瓶}
        \]
        1000 瓶：
        \[
        1000 \times 0.1063 = 106.3\,\text{元}
        \]
        \end{tcolorbox}
    \end{minipage}
\end{minipage}

\noindent\rule{\textwidth}{0.5pt}

\noindent
\begin{minipage}{\textwidth}
    \begin{minipage}{0.3\textwidth}
        \centering
        \includegraphics[width=0.9\linewidth]{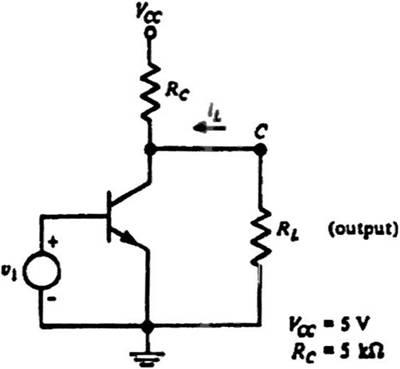}
        \captionof{figure}{example 3}
        \label{fig:inverter}
    \end{minipage}
    \hfill
    \begin{minipage}{0.65\textwidth}
        \begin{tcolorbox}[title=Prompt]
        The circuit in \textless image 1\textgreater{} consists of a general inverter connected to a load. Suppose the range of logical 1 is defined to be 4 to 5\,V. Determine the minimum $R_L$ in order that the output be 1 when the input voltage $v_1$ is zero.
        \end{tcolorbox}

        \begin{tcolorbox}[title=Model Response]
        \verb|<think>|...\verb|</think>|\\
        To determine the minimum $R_L$ so that $V_{out}$ stays between 4\,V and 5\,V when $v_1=0$\,V, note that the transistor is off, and the divider between $R_C$ and $R_L$ sets $V_{out}$:
        \[
        \frac{V_{cc} R_L}{R_C + R_L} \ge 4\,\text{V}
        \]
        with $V_{cc}=5$\,V and $R_C=5\,k\Omega$:
        \[
        \frac{5 R_L}{5\,k + R_L} \ge 4
        \]
        cross-multiplying:
        \[
        5 R_L \ge 4(5\,k + R_L) \implies R_L \ge 20\,k\Omega
        \]
        \textbf{Final Answer:}
        \[
        \boxed{20\,k\Omega}
        \]
        \end{tcolorbox}
    \end{minipage}
\end{minipage}

\begin{figure}[H]
    \centering
    \includegraphics[width=0.4\textwidth]{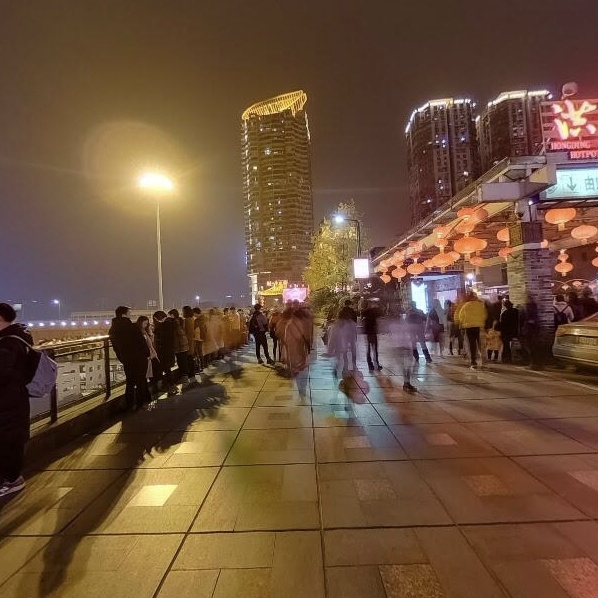}
    \caption{example 4}
    \label{fig:example2}
\end{figure}

\begin{tcolorbox}[title=Prompt]
You are participating in a geo-location challenge. Based on the provided image:

\begin{enumerate}
    \item Carefully analyze the clues in the image (including architectural style, signage, vegetation, terrain, etc.);
    \item Reason step by step to determine which continent, country, and city this location is most likely in, and explain your reasoning;
    \item Based on your analysis, estimate the approximate latitude and longitude.
\end{enumerate}

Please reason carefully and provide supporting evidence. Your final answer \textbf{must} include the following five lines of information, each on a separate line:

\begin{verbatim}
continent: [continent name]
country: [country name]
city: [city name]
lat: [latitude in decimal form]
lng: [longitude in decimal form]
\end{verbatim}
\end{tcolorbox}

\begin{tcolorbox}[title=Model Response]
\verb|<think>|...\verb|</think>|\\
\texttt{continent: Asia} \\
\texttt{country: China} \\
\texttt{city: Chongqing} \\
\texttt{lat: 29.5667} \\
\texttt{lng: 106.5500}

\vspace{0.5em}
\textbf{Step-by-Step Explanation:}

\begin{itemize}
    \item \textbf{Architectural Clues:} The tall building with a golden crown-like structure and the presence of red lanterns suggest a Chinese urban setting, indicative of modern Chinese architecture often found in cities like Chongqing.
    \item \textbf{Signage Analysis:} The visible ``WELCOME'' sign with Chinese characters, along with the bustling waterfront promenade, aligns with Chongqing’s famous riverside areas such as Nanbin Road.
    \item \textbf{Cultural Context:} Red lanterns and the crowded nighttime scene are characteristic of Chongqing’s vibrant urban culture and popularity as a tourist destination.
    \item \textbf{Geographical Indicators:} The hilly terrain and riverfront setting are hallmark features of Chongqing’s topography.
\end{itemize}

Thus, combining these observations, the location is most likely in Chongqing, China.
\end{tcolorbox}

\end{CJK}

\end{document}